\documentclass{article}

\usepackage{microtype}
\usepackage{graphicx}
\usepackage{subcaption}
\usepackage{booktabs} % for professional tables

\usepackage{hyperref}
\usepackage{stfloats}
\usepackage{booktabs}
\usepackage{float}
\usepackage{appendix}
\usepackage[dvipsnames,svgnames,table]{xcolor}

\definecolor{figBlue}{HTML}{3498DB}   % A medium-light blue match
\definecolor{figRed}{HTML}{E74C3C}    % A slightly orangey-red match for point A
\definecolor{figOrange}{HTML}{F39C12} % A standard flat orange match

\newcommand{\figblue}[1]{{\textcolor{figBlue}{#1}}}
\newcommand{\figred}[1]{{\textcolor{figRed}{#1}}}
\newcommand{\figorange}[1]{{\textcolor{figOrange}{#1}}}
\newcommand{\mbf}[1]{\mathbf{#1}}
\usepackage[accepted]{icml2026}

\usepackage{amsmath}
\usepackage{amssymb}
\usepackage{mathtools}
\usepackage{amsthm}

\usepackage[capitalize,noabbrev]{cleveref}

\theoremstyle{plain}
\newtheorem{theorem}{Theorem}[section]

\theoremstyle{definition}
\newtheorem{definition}[theorem]{Definition}

\theoremstyle{remark}

\usepackage[textsize=tiny]{todonotes}

\icmltitlerunning{Multi-Objective Bayesian Optimization via Adaptive $\varepsilon$-Constraints Decomposition}
 
\begin{document}

\twocolumn[
  \icmltitle{Multi-Objective Bayesian Optimization via Adaptive $\varepsilon$-Constraints Decomposition}



  \icmlsetsymbol{equal}{*}

    \begin{icmlauthorlist}
    \icmlauthor{Yaohong Yang}{aalto,*}
    \icmlauthor{Sammie Katt}{aalto,*}
    \icmlauthor{Samuel Kaski}{aalto,*,manchester}
  \end{icmlauthorlist}

  \icmlaffiliation{aalto}{Department of Computer Science, Aalto University, Espoo, Finland}
  \icmlaffiliation{*}{ELLIS Institute Finland}
  \icmlaffiliation{manchester}{Department of Computer Science, University of Manchester, Manchester, United Kingdom}
  \icmlcorrespondingauthor{Yaohong Yang}{yaohong.yang@aalto.fi}

  \icmlkeywords{Machine Learning, ICML}

  \vskip 0.3in
]



\printAffiliationsAndNotice{}  % no special notice (required even if empty)

\begin{abstract}

    Multi-objective Bayesian optimization (MOBO) provides a principled framework for optimizing multiple expensive black-box functions. 
    However, existing MOBO methods often struggle with coverage, scalability, and handling constraints and preferences.
    In this work we propose \textit{STAGE-BO, Sequential Targeting Adaptive Gap-Filling $\varepsilon$-Constraint Bayesian Optimization}:
    by analyzing the coverage of the surrogate Pareto front, our method identifies the Pareto front point with the largest uncovered gap, and uses its coordinates to define adaptive constraints in $\varepsilon$-constraint method, which transforms the problem into a sequence of inequality-constrained subproblems, efficiently solved via constrained expected improvement acquisition. 
    Our approach provides uniform Pareto coverage without hypervolume computation and naturally handles constraints and preferences. 
    Experiments on synthetic and real-world benchmarks demonstrate superior coverage and competitive hypervolume performance against state-of-the-art baselines.
    Our code implementation can be found at \url{https://github.com/YangYaohong1/STAGE-BO}.
    % In practice, however, existing MOBO methods face several challenges: scalarization-based methods often fail to capture diverse trade-offs, hypervolume-based approaches are computationally expensive, and incorporating constraints or user preferences typically requires substantial algorithmic modifications.
    % In this work, we propose Adaptive Gap-Filling $\varepsilon$-Constraint Bayesian Optimization, \red{a unified MOBO framework} that explicitly targets under-explored regions of the Pareto front while naturally accommodating constraints and preference information. 
    % Building on the $\varepsilon$-constraint formulation, our method adaptively selects constraint thresholds based on the geometry of the posterior Pareto front, identifying and filling the largest geometric gaps among previously evaluated solutions. 
    % This transforms the original multi-objective problem into a sequence of inequality-constrained single-objective subproblems, which we solve efficiently using constrained Thompson sampling and constrained expected improvement strategy.
    % Our approach systematically promotes diverse Pareto-optimal solutions without relying on hypervolume computation and applies seamlessly to unconstrained, constrained, and preference-based MOBO settings. 
    % Empirical results on synthetic benchmarks and real-world tasks demonstrate that the proposed method achieves improved Pareto front coverage and competitive optimization performance compared to state-of-the-art MOBO baselines.
    
\end{abstract}

\section{Introduction} \label{sec:intro}

    Multi-objective Bayesian optimization (MOBO) has emerged as a powerful paradigm for optimizing multiple expensive black-box functions \cite{daulton2020differentiable,tu2022joint,ngomobo}. 
    % By extending standard Bayesian optimization \cite{garnett2023bayesian} to the multi-objective setting,
    By navigating trade-offs, MOBO facilitates discovery in diverse fields ranging from machine learning \cite{sener2018multi}, materials science \cite{xu2025multi} to robotics \cite{kouritem2022multi}. 
    % In these domains, improving one objective often necessitates compromising another; thus, the goal is not to identify a single optimum, but to approximate the set of \textit{Pareto-optimal solutions}.
    % MOBO extends Bayesian Optimization (BO) \cite{garnett2023bayesian} -- a powerful framework for optimizing expensive black-box functions -- to settings where multiple, often conflicting, objectives must be optimized simultaneously. 
    % Applications of MOBO include but are not limited to machine learning \cite{sener2018multi,karl2023multi}, material design \cite{deshwal2021bayesian, xu2025multi}, agriculture \cite{jiang2020multi, karamian2023application}, robotics \cite{kouritem2022multi, wei2020particle} and vehicle design \cite{kohira2018proposal, anosri2023comparative}.

    % intro of MOBO 
    % copy from MOBO-OSD
    \begin{figure}[ht]  
      \vskip -0.1in
      \begin{center}
        \centerline{\includegraphics[width=\columnwidth]{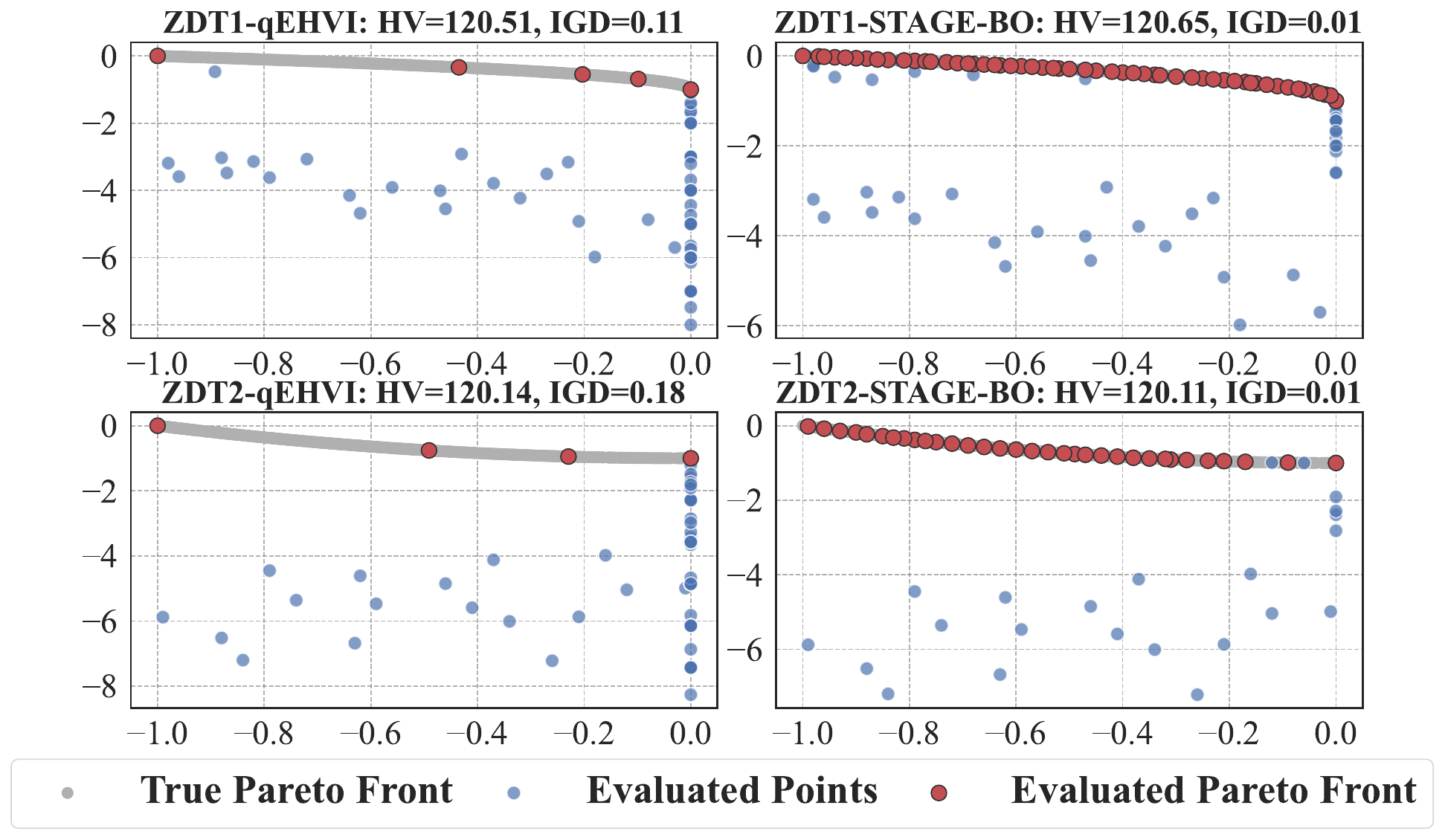}}
        \caption{
        \textbf{Pareto front approximation on ZDT1 and ZDT2 benchmarks.}  While qEHVI (left) and our method (right) achieve comparable hypervolume (HV), ours yields an order-of-magnitude reduction in inverse generational distance (IGD), demonstrating significantly better uniform coverage of the Pareto front. This motivates the need for multiple metrics to assess solution quality.
        % This discrepancy demonstrates highlights the necessity of reporting complementary metrics like IGD besides hypervolume to fully characterize the geometric quality of the approximation.
        }
        \label{fig:motivation}
        \vskip -0.2in
      \end{center}
    \end{figure}   
    % the bias of maximizing HV
    The objective functions in MOBO are often conflicting, meaning that improving one objective may deteriorate another. 
    Hence the goal is not to identify a single optimal solution, but rather a set of \textit{Pareto optimal solutions.}
    % The most popular and commonly used evaluation metric is hypervolume (HV), which is to compute the area covered by the Pareto front to a reference point. 
    % It is widely regarded as the standard metric, largely because it is the only strictly Pareto-compliant unary metric. 
    % Therefore, most algorithms aim to maximize HV. 
    The standard metric for evaluating Pareto approximations is hypervolume (HV), which measures the size of the dominated objective space.
    {Because HV is strictly Pareto-compliant}, the majority of MOBO algorithms aim to maximize HV improvement.
    However, this reliance on HV introduces two limitations. 
    First, \textbf{computational scalability:} exact HV computation scales exponentially with the number of objectives, making it prohibitively expensive for the problem with more than four objectives. 
    Second, \textbf{{geometric} bias:} analysis by \citet{auger2009theory} shows that asymptotically, the density of solutions maximizing HV is proportional to the square root of the Pareto front's negative slope ($\propto \sqrt{-F'(\mbf{x})}$, where $F=[f_1,\dots, f_m]$) are the objectives. 
    Consequently, HV maximization exhibits a bias: it heavily concentrates solutions in steep \textit{knee} regions while undersampling \textit{flat} trade-off areas, where the slope is small or close to zero, failing to uniformly cover the front.
    
    Empirical evidence supports this theoretical concern.
    As illustrated in \cref{fig:motivation}, we observe that while qEHVI \cite{daulton2020differentiable} achieves high HV scores, it fails to generate a uniformly diverse set of solutions.
    Thus its inverse generational distance (IGD)---the distance from the true Pareto front to the obtained Pareto front---can be an order of magnitude larger than that of our proposed method.
    This discrepancy underscores the limitation of relying solely on HV for evaluation and highlights the necessity of using multiple metrics, such as HV and IGD, to comprehensively assess Pareto front quality in the context of algorithm design.
    % taking hypervolume maximization as the acquisition function can effectively find points that return high hypervolume, but fail to generate a set of diverse Pareto optimal points, which is bad for the decision makers to do the downstream tasks. 
    % Though EHVI can achieve high hypervolume, the IGD is ten times larger than our method. 
    % Therefore, we propose that it is better to use multiple metrics to measure the quality of Pareto front and compare different methods. 
    
    Diversity-guided MOBO methods attempt to improve coverage but often face significant bottlenecks: they either rely on {input-space metrics }\cite{renganathan2025q} that do not guarantee diversity in the objective space, or retain the computational cost of HV maximization \cite{konakovic2020diversity, ngomobo, ahmadianshalchi2024pareto}.
    Scalarization methods \cite{paria2020flexible,knowles2006parego} avoid these costs by decomposing the problem via random weights. 
    However, it is well-established that a uniform distribution of weights does not map to a uniform set of solutions on the Pareto front \cite{das1998normal}, frequently resulting in clustered solutions and large geometric gaps.

    In this work, we propose {\textit{Sequential Targeting via Adaptive Gap-Filling $\varepsilon$-Constraint Bayesian Optimization (STAGE-BO)}} to efficiently generate uniform Pareto optimal solutions without the biases or costs of hypervolume computation.
    We build upon the key insight of the $\varepsilon$-constraint method \cite{haimes1971bicriterion, chankong2008multiobjective, branke2008multiobjective}: any Pareto-optimal solution can be recovered by optimizing one objective while constraining the others \cite{mavrotas2009effective}.
    % By turning MOO into a sequence of single-objective problems, $\varepsilon$-constraint method directly encodes acceptable trade-offs, makes the method naturally interpretable, and aligns well with user preferences. 
    % The practical challenges are that choosing $\varepsilon$ values is difficult. 
    % Choosing $\varepsilon$ uniformly in objective space may lead to infeasible subproblems or clustered solutions.
    % Unlike weight-based scalarization, this formulation directly encodes acceptable trade-offs and is robust to non-convexity. 
    % The practical challenge, however, lies in selecting the constraint thresholds $\varepsilon$; choosing them naïvely can lead to infeasible subproblems or redundant sampling.
    The practical challenge lies in selecting constraint thresholds $\varepsilon$ that leads to uniform coverage of the Pareto front.
    Our method solves this issue by identifying the largest gap between the surrogate Pareto front and the observations, which are then set as constraints in the $\varepsilon$-constraint method.
    This way, we transform the global MOBO problem into a sequence of inequality-constrained sub-problems solved via constrained expected improvement \cite{schonlau1998global, gardner2014bayesian}. 
    % Since we directly operate on the objective space, this approach naturally unifies unconstrained, constrained, and preference-aware MOBO into a framework.

    Our specific contributions are as follows: 
    \begin{itemize} 
    \item We propose STAGE-BO: an adaptive gap-filling $\varepsilon$-constraint method that selects constraints based on the geometry of the surrogate Pareto front (measured by \emph{fill distance}), thereby guaranteeing uniform coverage without hypervolume computation. 
    \item We introduce a general framework that seamlessly handles standard MOO, constrained MOO, and preference-aware MOO without requiring structural modifications.
    \item We demonstrate through extensive experiments that STAGE-BO achieves superior Pareto coverage and competitive hypervolume performance compared to state-of-the-art baselines on both synthetic benchmarks and real-world tasks. 
    \end{itemize}
    % Our method adaptively selects constraint thresholds based on posterior Pareto front geometry.
    % Specifically, it dynamically identifies the largest geometric voids (gaps) between the posterior Pareto front and the evaluated solutions and place the constraints, then transforms the global optimization problem into a series of inequality-constrained sub-problems. 
    % To solve these sub-problems efficiently, we employ the constrained expected improvement as the acquisition function.  
    % Moreover, we further show that our formulation naturally handles MOBO with unknown constraints and users' preferences without structural modification, to demonstrate the flexibility of our method. 
    % Empirically, we demonstrate that our method achieves superior sample efficiency and diversity compared to state-of-the-art baselines across a suite of benchmark problems and real world applications. 
    % We further show that our formulation naturally handles constrained optimization and user-defined regions of interest (ROI)

    % Empirically, we demonstrate that our framework offers a flexible trade-off between computational cost and sample efficiency. 
    % The Thompson Sampling variant reduces solver overhead by orders of magnitude compared to qNEHVI, while the cEI variant achieves state-of-the-art coverage on constrained and ROI-based tasks.

\section{Background} \label{sec:background}
\subsection{Bayesian Optimization} \label{subsec:bo}
    
    Bayesian Optimization (BO) \cite{garnett2023bayesian} is a sequential design strategy for the global optimization of black-box functions that are expensive to evaluate. 
    Formally, we seek to find a global maximizer $\mbf{x}^*$ of an objective function: $f:\mbf{\mathcal{X}}\rightarrow\mathbb{R}$ over a bounded domain $\mathcal{X}\subset\mathbb{R}^d$:
    $
        \mbf{x}^*=\arg\max_{\mbf{x}\in\mathcal{X}}f(\mbf{x}).
    $
    The BO framework rests on two principal components: a probabilistic surrogate model and an acquisition function. 
    A Gaussian Process (GP) \cite{williams2006gaussian} is typically employed as a prior distribution over $f(\mbf{x})$. 
    The GP is fully specified by a mean function and a covariance kernel, denoted as $f(\mbf{x})\sim \mathcal{GP}(m(\mbf{x}), k(\mbf{x},\mbf{x}'))$.
    Given a dataset of observations $\mathcal{D}_t=\{(\mbf{x}_i, \mbf{y}_i\}_{i=1}^t$, 
    % the posterior distribution provides a predictive mean $\mu_t(\mbf{x})$ and variance $\sigma^2_t(\mbf{x})$ for an unobserved point $\mbf{x}$.
    an acquisition function $\alpha(\mbf{x}|\mathcal{D}_t)$ is maximized to sample the next query: $\mbf{x}_{t+1}=\arg\max_{\mbf{x}\in\mathcal{X}} \alpha(\mbf{x}|\mathcal{D}_t)$. 
    The objective $f$ is then evaluated at $\mbf{x}_{t+1}$, the dataset is updated 
    % to $\mathcal{D}_{t+1}=\mathcal{D}_{t}\cup\{(\mbf{x}_{t+1},\mbf{y}_{t+1})\}$, 
    and the posterior is recomputed. 
    % This process repeats until a convergence criterion or a budget constraint is met. 
    % from mobo-osd 
    % Bayesian Optimization (BO) \cite{garnett2023bayesian} is a common tool for optimizing expensive black-box objective functions $f$. 
    % Given a minimization problem, the goal is to find the global optimum of the function $f$ using the least function evaluations. 
    % BO sequentially selects observation data via an iterative process. 
    % Each BO iteration trains a probabilistic surrogate model, builds an acquisition function, then selects the acquisition function's maximizer as the next observation. 
    % The most common type of surrogate model for BO is a Gaussian Process (GP) \cite{williams2006gaussian}, which provides a posterior distribution over the objective function given the observed dataset $\mathcal{D}$. 
    % The acquisition function $\alpha:\mathcal{X}\rightarrow\mathbb{R}$ is constructed from the surrogate model and an optimization policy to quantify the utility of each unobserved data point. 
    The common acquisition functions are EI \cite{movckus1974bayesian}, UCB \cite{srinivas2010gaussian} and TS \cite{thompson1933likelihood}.

\subsection{Multi-Objective Optimization} \label{subsec:background_mobo}
    A multi-objective optimization (MOO) problem has a vector-valued objective function $F: \mathcal{X}\rightarrow\mathcal{Y}$ with $F=(f_1,\dots,f_m),$ where $\mathcal{X}\in \mathbb{R}^d$ is a $d$-dimensional input space, and $\mathcal{Y}\in\mathbb{R}^m$ is an $m$-dimensional output space ($m>1)$.
    Without loss of generality, we assume the problem is to maximize all objectives of $F$:  $\max_{\bf{x}\in\mathcal{X}}\; F(\mbf{x})= [f_1(\mbf{x}),\dots,f_m(\mbf{x})].$
    In MOO, the goal is to identify the set of \textit{Pareto optimal solutions}, all of which are mathematically equivalent when no preference information is specified.
    
    \paragraph{Pareto Optimality} 
    For a pair $(\mbf{x}, \mbf{x}')$,  we say ``$\mbf{x}$ weakly dominates $\mbf{x}'$'' if $F(\mbf{x})$ is no worse than $F(\mbf{x}')$ in all objectives, i.e. $f_i(\mbf{x}) \geq f_i(\mbf{x}')$ for all $i\in \{1,\dots,m\}$. 
    If at least one of the inequalities is strict, we say ``$\mbf{x}$ dominates $\mbf{x}'$''. 
    If $\mbf{x}$ is not (weakly) dominated by any other $\mbf{x}'$, $\mbf{x}$ is called (weakly) Pareto-optimal. 
    \emph{(Weak) Pareto front} $\mathcal{P}_f$ is a set of (weakly) Pareto optimal solutions, and the corresponding set of Pareto optimal inputs is called the \emph{Pareto set} $\mathcal{P}_s$.

    Multi-Objective Bayesian Optimization (MOBO) extends BO to optimize expensive black-box, vector-valued objective functions $F$. 
    Given a maximization problem, the goal is to identify the Pareto set $\mathcal{P}_s$ and the corresponding Pareto front $\mathcal{P}_f$, using a minimal number of function evaluations. 
    MOBO frameworks typically employ independent GPs to model each objective function $f_i$ \cite{ bradford2018efficient,paria2020flexible,belakaria2020uncertainty,daulton2020differentiable}. 
    An acquisition function is optimized to select the next query.
    See the detailed discussion in \Cref{sec:related_work}.
    % For acquisition functions, \red{while several works leverage existing single-objective acquisition functions }\cite{knowles2006parego, paria2020flexible,belakaria2020uncertainty}, others propose new acquisition functions tailored for the MOO setting \cite{hernandez2016predictive,belakaria2019max,tu2022joint,daulton2020differentiable,daulton2021parallel,daulton2023hypervolume}.
    
    Two commonly used performance metrics in MOO are hypervolume and inverted generational distance.
    \textit{Hypervolume} (HV) \cite{zitzler1998multiobjective} is defined as the $m$-dimensional Lebesgue measure $\lambda_m$ of the space dominated by solutions and bounded by the reference point $\mbf{r}$:
    \begin{equation}\label{eq:hv}
        \text{HV}(\mbf{Y},\mbf{r})=\lambda_m(\cup_{\mbf{y}\in\mbf{Y}}[\mbf{r}, \mbf{y}]),
    \end{equation}
    where $[\mbf{r},\mbf{y}]$ denotes the hyperrectangle bounded by the reference point $\mbf{r}$ and solutions $\mbf{y}\in\mbf{Y}$.
    
    \textit{Inverted generational distance} (IGD) \cite{coello2004study} assesses both the convergence and diversity of the approximation by measuring the average distance from the true Pareto front $\mathcal{P}_f$ to the nearest solution in the observed set $\mathbf{Y}$: 
    % \todo{we need to make sure that readers understand this is an important measure of diversity. We do not need to say that specifically here, it is just important that it is done at some point.}
    \begin{equation} \label{eq:igd}
        \text{IGD}(\mbf{Y}, \mathcal{P}_f) =\frac{1}{|\mathcal{P}_f|}(\sum_{\mbf{y}'\in\mathcal{P}_f}\min_{\mbf{y}\in\mathbf{Y}}\|\mbf{y}-\mbf{y}'\|).       
    \end{equation}
    Low IGD values indicate that the front approximation is both close to the true front and covers it uniformly.
           
\subsection{Constrained Multi-Objective Optimization} \label{subsec:background-conmobo}
    In practical engineering and scientific applications, valid solutions often satisfy safety or physical constraints alongside objective trade-offs \cite{fromer2023computer, gardner2019constrained}.
    The constrained multi-objective optimization problem is defined as:
    \begin{align}
        \max_{\mbf{x}\in \mathcal{X}}\; &F(\mbf{x})\;=\; [f_1(\mbf{x}),\dots,f_m(\mbf{x})] \notag \\ 
        s.t.\; &G(\mbf{x})\; =\;[g_1(\mbf{x}),\dots,g_q(\mbf{x})]\geq 0,
    \end{align}
    where $G(\mbf{x}):\mathcal{X}\rightarrow\mathcal{C}\in \mathbb{R}^q$ is the $q$ constraints functions.
    Both $F(\mbf{x})$ and $G(\mbf{x})$ are unknown.  
    Consequently, the search space is restricted to the feasible region given by 
    \begin{equation} \label{eq:feasible_region}
        \mathcal{Q} = \{\mbf{x}|\mbf{x}\in \mathcal{X}, g_l(\mbf{x})\geq 0,\forall l\in[q] \}.
    \end{equation}

    The goal is to identify the Pareto front $\mathcal{P}_f$ and Pareto set $\mathcal{P}_s$ strictly within $\mathcal{Q}$.
    This setting introduces complexity, as the optimizer must simultaneously learn the boundaries of the feasible region and maximize the objectives.

     A standard approach to constrained MOBO is model each objective $f_i(\mbf{x})$ and constraint $g_j(\mbf{x})$ using independent GPs. 
     The search for feasible Pareto front is guided by modifying the acquisition function to account for constraint satisfaction, such as cEHVI \cite{abdolshah2018expected}, qPOTS \cite{renganathan2025q}, COMBOO \cite{li2025constrained}.

\subsection{Preference-Aware Multi-Objective Optimization} \label{subsec:background-premobo}

    While standard MOO aims to approximate the entire Pareto front, in many decision-making scenarios, the whole front is computationally expensive to recover \cite{paria2020flexible, chen2024mosh}.
    Instead, decision-makers often possess prior knowledge regarding acceptable trade-offs. 
    This motivates the preference-aware setting, where the goal is to concentrate evaluations solely on a user-defined Region of Interest (ROI).
    % ; however, the feasible region may be empty, since users may not know how to specify constraints that cover the Pareto front or the feasible region.
    Following \citet{paria2020flexible, hakanen2017using}, we adopt the bounding box formulation, where preferences are expressed as thresholds. 
    The ROI is defined as $\mathcal{B}=\{\mbf{y}\in\mathbb{R}^m|a_i
    \leq \mbf{y}\leq b_i,\forall i=1,\dots,m\}$. 
    The optimization goal is to recover the Pareto optimal solutions satisfying $F(\mathbf{x}) \in \mathcal{B}$.
    % \citet{paria2020flexible} convert bounding boxes to a weight distribution, turn the multi-obejctive problem into a single objective problem and employ acquisition functions such as TS and UCB.

\section{Related Work} \label{sec:related_work}
    % \paragraph{Bayesian Optimization} 
    % % copy from pareto set learning...
    % Surrogate model-based methods have been widely used and studied for expensive optimization [47, 40, 65, 79]. 
    % These methods iteratively build a surrogate model to approximate the black-box objective function, and uses an acquisition function to search for the optimal solution. 
    % Much effort has been made on various design issues in Bayesian optimization, such as acquisition functions [81], high-dimensional optimization [91, 92], batch evaluation [22], scalable optimization [80, 27], and theoretical analysis [42]. 
    % Most work for BO are on single-objective optimization. We refer readers to Garnett [30] for a comprehensive introduction.

    \paragraph{Multi-Objective Bayesian Optimization}
    MOBO methods largely fall into three categories: scalarization, hypervolume (HV) maximization, and information-theoretic approaches. 
    Scalarization methods (ParEGO \cite{knowles2006parego} and TS-TCH \cite{paria2020flexible}) decompose the problem into single-objective subtasks using random weights, which often fail to cover fronts uniformly. 
    HV-based methods (EHVI \cite{emmerich2008computation}, qEHVI \cite{daulton2020differentiable}, and TSEMO \cite{bradford2018efficient}), prioritize maximizing the dominated volume.
    As discussed in \Cref{sec:intro}, they suffer from the intrinsic bias and high computational costs that scale exponentially with the number of objectives. 
    Information-theoretic methods (PESMO \cite{hernandez2016predictive}, MESMO \cite{belakaria2019max}, PFES \cite{suzuki2020multi}, and JESMO \cite{tu2022joint}) maximize information gain about the Pareto front but often require heavy approximations to compute.

    Recent work attempts to explicitly enforce coverage but typically retains bottlenecks.
    DGEMO \cite{konakovic2020diversity} guides the search toward diverse regions but still employs HV improvement for the final selection.
    Moreover, its reliance on data structures limits the scalability beyond three objectives. 
    PDBO \cite{ahmadianshalchi2024pareto} employs a bandit strategy with determinantal point processes to select diverse batches. 
    MOBO-OSD \cite{ngomobo} decomposes the problem via orthogonal search directions but still relies on HV maximization for selection. 
    qPOTS \cite{renganathan2025q} combines Thompson Sampling with a maximin strategy; however, it calculates diversity in the input space, which does not guarantee output space diversity. 
    Unlike these approaches, our method ensures output-space diversity without calculating HV.

    \paragraph{Constrained Multi-Objective Bayesian Optimization} 
    % copy from COMBOO
    Research explicitly targeting constrained MOBO remains relatively sparse.
    Standard frameworks, such as the widely used BoTorch \cite{balandat2020botorch} implementations of qEHVI \cite{daulton2020differentiable, daulton2021parallel}, typically handle constraints by weighting the primary acquisition value by the probability of feasibility, adopting the strategy originally proposed by \citet{gelbart2014bayesian}.
    Within the information-theoretic paradigm, \citet{hernandez2016predictive} extended Predictive Entropy Search (PES) to constrained settings, explicitly balancing constraint learning with Pareto frontier discovery. 
    To mitigate the high computational cost of PES, \citet{fernandez2023improved} subsequently proposed a formulation based on Max-value Entropy Search (MES) \cite{wang2017max}.
    However, these entropy-based methods rely on heavy approximations to maintain tractability, which can degrade performance in complex landscapes. 
    Most recently, qPOTS \cite{renganathan2025q} can also handle constraints naturally.
    COMBOO \cite{li2025constrained} introduced a scalarization-based approach that combines random weights with optimistic feasibility assessments (UCB) to identify the constrained Pareto front.

    \paragraph{Preference-Aware Multi-Objective Bayesian Optimization}
    % copy from ozaki
    The integration of Decision Maker preferences has been stuided in MOO.
    Early approaches \cite{abdolshah2019multi} encoded these preferences via objective importance rankings.
    However, a more prevalent formulation defines the preference structure as a specific Region of Interest (ROI) in the objective space, typically bounded by reference vectors or hyper-rectangles \cite{hakanen2017using, paria2020flexible, palar2018multi, he2020preference}.
    % In particular, \citet{paria2020flexible} systematically generalize the random scalarization technique to consider the user's preferred region. 
    \citet{paria2020flexible} systematically generalize the random scalarization technique, allowing different scalarization techniques, e.g., weighted sum and Tchebyshev \cite{nakayama2009sequential}, as well as different acquisition functions, e.g., TS \cite{thompson1933likelihood}, UCB \cite{srinivas2010gaussian} to concentrate candidate generation specifically within the user's preferred region.
    % To avoid generating weakly Pareto-optimal solutions—points that satisfy constraints but are dominated in the constraint objectives—

\section{Methodology} \label{sec:method}

    % Our key motivation is the observation that optimizing hypervolume -- the typical metric of interest in MOO -- may unintentionally lead to poor coverage of the Pareto front.
    % This work aims to remedy this by identifying potential gaps in the current observed Pareto front.
    % Our method consists of (iterative application of) two steps: first, it identifies objective values that are likely to be on the Pareto front according to our posterior, and not covered yet.
    % Second, it considers those objective values (all-but-one) as constraints and finds the query that optimally improves the Pareto front under those constraints.
    % Next (\cref{sec:proposed-method}), we will describe both steps in detail.
    % Then, in~\cref{sec:constrained-extension,sec:preference-extension}, we discuss extensions to respectively constrained and preference-based MOO.
    Our key motivation is the observation that optimizing hypervolume may lead to non-uniformly distributed Pareto optimal solutions and is computationally expensive.
    This work aims to generate a set of Pareto optimal solutions that uniformly cover $\mathcal{P}_f$.
    We exploit the property that the $\varepsilon$-constraint method can find any Pareto-optimal point with the appropriate constraints on the objectives.
    Thus, the challenge is finding the constraints such that the queries found by the $\varepsilon$-constraint method lead to uniform coverage.
    Our method, called STAGE-BO, tackles this by identifying the largest under-explored region (gap) in the objective space.
    % To achieve this efficiently, we propose STAGE-BO, a framework that iteratively identifies the largest under-explored region (gap) in the objective space and targets it explicitly with $\varepsilon$-constraint method.

    To measure and, ultimately, optimize for uniform coverage, we adopt the \textit{Fill Distance} (FD) metric defined in \citet{zhang2024gliding}.
    For $\mathcal{D}_t=\{\mathbf{X}_t,\mathbf{Y}_t\}$, 
    \begin{definition}\label{eq:df}  
       $ \text{FD}(\mbf{Y}_t) =\max_{\mbf{y}\in\mathcal{P}_f}\min_{\mbf{y}'\in\mbf{Y}_t}\|\mbf{y}-\mbf{y}'\|,$
    \end{definition}  
    where $\|\cdot\|$ denotes the Euclidean distance between two points.
     \citet{zhang2024gliding} establish the relationship between FD and IGD and prove that the optimal FD configuration sets an upper bound for the IGD value.
    \begin{theorem}[\citet{zhang2024gliding}]
        Assume the goal is to find a set of Pareto optimal solutions, i.e., $\mbf{Y}\subset \mathcal{P}_f$ to optimize either the FD or IGD indicator: $\min_{\mbf{Y}\subset \mathcal{P}_f}\text{FD}(\mbf{Y})$ or $\min_{\mbf{Y}\subset \mathcal{P}_f}\text{IGD}(\mbf{Y})$ to reach a diverse distribution. 
        Let the optimal sets be $\mbf{Y}^{\text{FD}}$ and $\mbf{Y}^{\text{IGD}}$ respectively. 
        Then 
        \begin{equation}\label{theorem:fd}
            \text{IGD}(\mbf{Y}^{\text{IGD}})\leq \text{IGD}(\mbf{Y}^{\text{FD}})\leq \text{FD}(\mbf{Y}^{\text{FD}}).
        \end{equation}
    \end{theorem}
    Since the optimal IGD configuration does not similarly bound FD, we focus on minimizing FD in this paper.

    % Our method builds on the $\varepsilon$-constraint method \cite{haimes1971bicriterion, chankong2008multiobjective} to minimize FD, which decomposes the multi-objective problem into a sequence of inequality-constrained single-objective subproblems:
    
\subsection{Proposed Method: STAGE-BO} \label{sec:proposed-method}

    STAGE-BO consists of four main steps:
    1) Sample the surrogate objective functions; 
    2) Identify the Pareto front point with the largest uncovered gap, quantified by FD;
    3) Translate the target's coordinates into constraints, formulating the MOO problem into a sequence of inequality-constrained subproblems;
    4) Solve the resulting constrained optimization problem.
    We present the details below.

    FD identifies locations that are not yet covered by observations.
    % Ideally, we identify the points on the true Pareto front that maximize the fill distance.
    Since the true Pareto front $\mathcal{P}_f$ is not known, we resort to sampling from the posterior.
    
    % Specifically, we generate the posterior Pareto front $\widetilde{\mathcal{P}}_f^t$ from Thompson sampling, compute the point on $\widetilde{\mathcal{P}}_f^t$ with the highest FD and place constraints there. 
    % We present the details as following. 

    % We propose an iterative framework (\Cref{fig:stage_bo_demo}) that explicitly \red{targets} the largest geometric gaps in the current Pareto approximation and place the constraints there.
    % We show the details in the following sections.

    \begin{figure}[t]
     \vskip -0.1in 
      \begin{center}
        \centerline{\includegraphics[width=0.7\columnwidth]{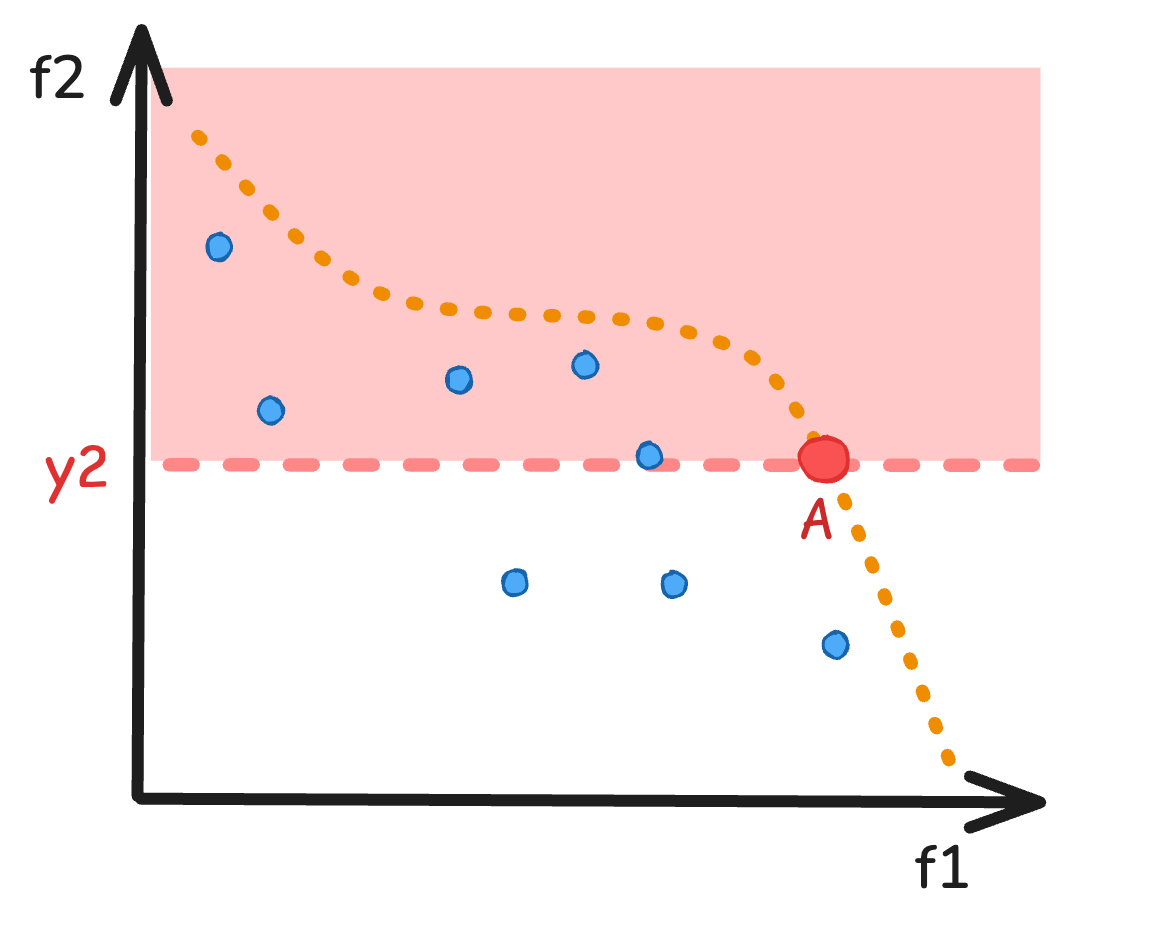}}
        \caption{Illustration of the STAGE-BO algorithm. The \figblue{blue dots} represent observations. The dashed \figorange{orange curve} depicts the sampled Pareto front approximation ($\widetilde{\mathcal{P}}_f^t$), generated via Thompson sampling and NSGA-II. The \figred{red point A $= (y_1, y_2)$} is identified on $\tilde{\mathcal{P}}_f^t$ as having the maxmin distance to the existing \figblue{blue observations}. Assuming a schedule where $f_1$ is targeted for optimization at this step, a constraint is established on the remaining objective using the objective value of $A$ ($f_2 \geq y_2$). The resulting valid search space for maximizing $f_1$ is indicated by the light \figred{red shaded region}.}
        \label{fig:stage_bo_demo}
      \end{center}
       \vskip -0.2in 
    \end{figure}

    \paragraph{Target Identification}
    We employ Thompson sampling \cite{wilson2020efficiently} to approximate the objectives. 
    Specifically, we draw a spectral sample path
    \begin{equation}\label{eq:ts-path}
        \tilde{F}^t(\mbf{x})=[\tilde{f}_1^t(\mbf{x}),\dots,\tilde{f}_m^t(\mbf{x})], \; \tilde{f}_i^t(\cdot)\sim \mathcal{GP}(f_i|\mathcal{D}_t)
    \end{equation}
    from the posterior distribution of the objective functions. 
    We then compute the \textit{sampled Pareto front} $\widetilde{\mathcal{P}}_f^t$ by maximizing this sampled trajectory:
    \begin{equation} \label{eq:nsga}
        \widetilde{\mathcal{P}}_f^t = \max_{\mbf{x}\in \mathcal{X}}\; [\tilde{f}_1(\mbf{x}),\dots,\tilde{f}_m(\mbf{x})].
    \end{equation}
    This cheap MOO problem is solved  using evolutionary algorithms such as NSGA-II \cite{deb2002fast}, generating a discrete approximation of the front.

    {Another option is to use the posterior mean as the surrogate path.} 
    However, as our ablation study in \Cref{app:ablation-ts} shows, replacing Thompson-sampled path with the posterior mean results in worse performance. 
    This is because the posterior mean is overly greedy and suppresses the uncertainty-driven variability needed for effective exploration.
    
    On the sampled front $\widetilde{\mathcal{P}}_f$, we seek the target $\mbf{Y}_c$ that approximates the center of the largest under-explored region from our current observations $\mbf{Y}_t$:
    \begin{equation} \label{eq:maxmin}
        \mbf{Y}_c = \arg\max_{\mbf{y}'\in \widetilde{\mathcal{P}}_f^t}\min_{\mbf{y}\in \mbf{Y}_t} \|\mbf{y}-\mbf{y}'\|.
    \end{equation}

    \paragraph{$\varepsilon$-constraint Decomposition}
    %The target $\mbf{Y}_c$ are objective values that, when seen as constraints for the $\varepsilon$-constraint method, leads to a query that maximizes the Pareto coverage.
    Having identified the target location $\mbf{Y}_c$, we require an optimization mechanism to guide the search toward it.
    To achieve this, we employ the $\varepsilon$-constraint method, which allows us to translate the coordinates of $\mathbf{Y}_c$ directly into search space boundaries.
    By setting the constraint thresholds based on the coordinates of $\mathbf{Y}_c$, we transform the multi-objective problem into a constrained single-objective subproblem whose unique optimal solution is guaranteed to be Pareto optimal \cite{branke2008multiobjective}. 
    %The $\varepsilon$-constraint method transforms the multi-objective problem into a constrained single-objective subproblem whose unique optimal solution is guaranteed to be Pareto optimal \cite{branke2008multiobjective}. 
    \begin{align} \label{eq:epsilon-constraint}
        &\max_{\mbf{x}\in\mathcal{X}} \; f_k(\mbf{x}) +s \sum_jf_j\notag \\
        & \text{subject to }f_j(\mathbf{x}) \geq \varepsilon_j\; \text{for all }j=1,\dots,m, j\neq k.
    \end{align}
    $s$ is set as a small number (e.g., $10^{-3}$) to avoid weakly Pareto optimal points.
    % Previous work finds the Pareto front by exhaustively iterating over all possible constraints.
    % We, instead, pick the constraints from $\mbf{Y}_c$.
    % It ensures the slack term acts only as a tie-breaker and never overpowers the primary objective $f_k$.
    % Under mild regularity assumptions, by varying $\varepsilon$ systematically, it can recover any Pareto-optimal solution \cite{branke2008multiobjective}. 
    This ensures that our next query is optimally positioned to fill the identified void.
    The detailed discussion can be found in \Cref{app:epsilon-constraioned-optimization}.
    
   % \todo{finish a very small but comprehensive motivation for what is happening below (i.e.``to guide the search'' is not saying why we are searching, or for what)}  
    % To guide the search toward $\mbf{Y}_c$, we employ the $\varepsilon$-constraint method to transform the MOO problem into a targeted subproblem.
    % We select one objective $f_k$ to optimize and treat the remaining $m-1$ objectives as constraints bounded by the coordinates of $\mathbf{Y}_c$.

    To avoid defining an empty feasible region, if $j$-th objective of the target component $\mathbf{Y}_{c,j}$ exceeds the maximum observed value for $j$-th objective, we clip the constraint to the best observed value:
    \begin{equation} \label{eq:relax}
        \widehat{\mbf{Y}}_{c,j}=
        \left\{\begin{matrix}
               \mbf{Y}_{c,j}  & \text{if }  \mbf{Y}_{c,j} <  \mbf{Y}_{t,j} \;\;\exists  t; \\ 
                \max\{\mbf{Y}_{t,j}\}_t   &  \text{if }  \mbf{Y}_{c,j} \geq \mbf{Y}_{t,j} \;\; \forall t.\\  
        \end{matrix}\right.
    \end{equation}

    {This clipping rule is designed as a numerical stabilizer:} it is triggered precisely when the target threshold exceeds all current observations on that objective, and usually happens in the early stage of the algorithm.
    An ablation study comparing STAGE-BO with and without clipping in \Cref{app:ablation-clip} shows that on most benchmarks the two variants perform comparably, confirming that clipping acts primarily as a stabilizer. 
    On a subset of benchmarks, clipping leads to measurable improvements, suggesting that the larger feasible region induced by clipping benefits optimization.
    
    To ensure balanced exploration across the objective space, we rotate the objective $f_k$ to be optimized in a round-robin fashion ($k = t \pmod m + 1$).
    This schedule guarantees that optimization pressure is distributed uniformly across all objectives over time.
    As demonstrated in \Cref{app:ablation-objective}, our framework is robust to the specific strategy used to select the objective for optimization.

    \paragraph{Acquisition Optimization}
    This decomposition transforms the original MOBO problem into a generic constrained Bayesian optimization problem.
    The constraints are placed at $m-1$ objectives $\{j|j\in (1,\dots,m), \text{ and } j\neq k\}$ with thresholds $\varepsilon_j=\widehat{\mathbf{Y}}_{c,j}$ and $k$-th objective is set to be optimized.
    We solve this efficiently using the Constrained Expected Improvement (cEI) acquisition function: \cite{schonlau1998global, gardner2014bayesian}.
    % \begin{enumerate}
    %     \item[(1)] cTS: 
    %     \begin{align}\label{eq:cts}
    %         &\mbf{x}_{t+1}=\arg\max_{\mbf{x}\in\mathcal{X}} \tilde{f}_{k}(\mbf{x})+s\sum_{j}f_j\notag \\ 
    %         &\text{subject to } \tilde{f}_j(\mbf{x})\geq \mathbf{Y}_c^j, j=1,\dots,m, j\neq k.
    %     \end{align}
        % Note we keep the same Thompson sampled curves as in the previous \Cref{eq:nsga}.
        % Therefore, the solution of \Cref{eq:cts} is exactly $\mbf{x}_c$. 
        % \todo{define $\mbf{x}_c$.}
        
        % Note qPOTS also uses Thompson sampling to sample curves, and use NSGAII to generate the posterior Pareto front. 
        % However, they compute the maxmin distance in the design space. 
        % We show in experiments section that our method improves dramatically over their method. 
        % This is because multi-objective functions are usually the `\textit{many-to-one}' problems, which means a small region in the design space might result in very close solutions.
        % Our method can improve the diversity in the objective space by directly considering the distance in the objective space. 
        
        \begin{equation}\label{eq:cEI}
            \mbf{x}_{t+1}
            % =\arg\max_{\mbf{x}\in\mathcal{X}}\alpha(\mbf{x})
            =\arg\max_{\mbf{x}\in\mathcal{X}}\text{EI}(\mbf{x})\times\text{PoF}(\mbf{x}),
        \end{equation}
        where $\text{EI}(\mbf{x})$ is the standard {Expected Improvement} of the objective $f_{k}$, and $\text{PoF}(\mbf{x})$ is the {Probability of Feasibility} satisfying the constraints.

        \begin{equation} \label{eq:ei}
            \text{EI}(\mbf{x})=\mathbb{E}[\max(0,f_{k}(\mbf{x})+s \sum_{j\neq k}f_j(\mbf{x})-f_{k}^* -s \sum_{j\neq k}f_j^*],
        \end{equation} 
        where $f_{k}^*+s \sum_{j\neq k}f_j^*$ is the best observed point.
        We assume the constraints are independent GPs. 
        \begin{equation}
            \text{PoF}(\mbf{x})=\prod_{j=1,\dots,m,j\neq k}\text{Pr}(f_j(\mbf{x})\geq \widehat{\mathbf{Y}}_{c,j}).
        \end{equation}

    By solving \Cref{eq:cEI}, STAGE-BO selects the sample that optimizes \Cref{eq:epsilon-constraint},
    effectively filling the identified gap.
    We visually show this process in \Cref{app:illustration}.

    \begin{algorithm}[H]
      \caption{The STAGE-BO Algorithm}
      \label{alg:stage-bo}
      \begin{algorithmic}[1]
        \STATE {\bfseries Input:} Evaluation budget $T$, initial dataset $\mathcal{D}_0$.
        \WHILE{$t \leq T$}
        \STATE Fit GP models on $\mathcal{D}_t$.
        \STATE Sample the posterior GP paths (\Cref{eq:ts-path}). 
        \STATE Solve the cheap MOO problems \Cref{eq:nsga} via NSGA-II.
        \STATE Identify the target point $\mbf{Y}_c$ on $\widetilde{\mathcal{P}}_f$ with the maxmin distance to evaluations $\mbf{Y}_t$ in \Cref{eq:maxmin}.
        \STATE Select primary objective: $k\leftarrow t\pmod{m}+1$.
        \STATE Set the constraints based on \Cref{eq:relax}.
        \STATE Optimize cEI (\Cref{eq:cEI}) to sample next point $(\mbf{x}_{t+1}, \mbf{y}_{t+1})$.
        \STATE Update the dateset $\mathcal{D}_{t+1}\leftarrow \mathcal{D}_{t}\cup (\mbf{x}_{t+1}, \mbf{y}_{t+1})$.
        \ENDWHILE
        \STATE{\bfseries Output:} The Pareto set $\mathcal{P}_s$ and Pareto front $\mathcal{P}_f$.
      \end{algorithmic}
    \end{algorithm}
    
\subsection{Extensions to constrained MOBO}  \label{sec:constrained-extension}
    % MOBO has been widely adopted in scientific experiment design, including drug discovery \cite{fromer2023computer} and hyperparameter optimization \cite{candelieri2024fair}. 
    In practice, regulatory or safety concerns often impose additional thresholds $G(\mathbf{x})$ on certain attributes of the experimental outcomes \cite{fromer2023computer}.
    STAGE-BO extends naturally to this setting without structural changes.

    Since our framework already reduces MOO to a sequence of constrained sub-problems (\Cref{eq:epsilon-constraint}), incorporating physical constraints is natural: we append the external constraints to the set of algorithmically generated $\varepsilon$-constraints. 
    The optimization problem becomes:
    \begin{align} \label{eq:constrained_epsilon-constraint}
        \max \; &f_k(\mbf{x}) +s \sum_jf_j\notag \\
         \text{subject to } &f_j(\mathbf{x}) \geq \varepsilon_j\; \text{for all }j=1,\dots,m, j\neq k \notag \\
        & g_l(\mbf{x}) \geq 0 \;\text{for all } l=1,\dots,q.
    \end{align}
    
    To ensure the identified target $\mathbf{Y}_c$ is physically reachable, we must also account for feasibility during the target identification phase.
    After collecting $t$ observations $\mathcal{D}_t=\{\mbf{X}_t,\mbf{Y}_t,\mbf{C}_t)\}=\{(\mbf{x}_i, \mbf{y}_i,\mbf{c}_i)\}_{i=1}^t$, we build GP models separately for each objective and each constraint.
    We sample paths for both objectives and constraints using Thompson sampling.
    Besides \Cref{eq:ts-path}, we also sample
    \begin{equation}\label{eq:con_ts-path}
        \tilde{G}^t(\mbf{x})=[\tilde{g}_1^t(\mbf{x}),\dots,\tilde{g}_c^t(\mbf{x})], \; \tilde{g}_i^t(\cdot)\sim \mathcal{GP}(g_i|\mathcal{D}_t).
    \end{equation}
    We optimize the following cheap constrained MOO problem to obtain the sampled Pareto front $\widetilde{\mathcal{P}}_f^t$ using NSGA-II:
    \begin{align} \label{eq:con_nsga}
        &\widetilde{\mathcal{P}}_f^t = \max_{\mbf{x}\in \mathcal{X}}\; [\tilde{f}_1(\mbf{x}),\dots,\tilde{f}_m(\mbf{x})], \notag \\
        &\text{subject to } \tilde{g}_l(\mbf{x})\geq 0,\; \forall l\in[q]
    \end{align}

    % We find the point with maxmin distance using \Cref{eq:maxmin} and \Cref{eq:relax} to decide to where to place the constraints.
    % Then cEI is adopted to solve \Cref{eq:constrained_epsilon-constraint}.
    This ensures that the target gap $\mathbf{Y}_c$ computed via \Cref{eq:maxmin} lies within the valid feasible region. 
    Finally, the acquisition optimization (\Cref{eq:cEI}) remains unchanged, with $\text{PoF}(\mathbf{x})$ updated to include the probability of satisfying physical constraints:
    \begin{equation}
    \text{PoF}(\mathbf{x}) = \prod_{j \neq k} \text{Pr}(f_j(\mathbf{x}) \geq \widehat{\mathbf{Y}}_{c,j}) \prod_{l=1}^q \text{Pr}(g_l(\mathbf{x}) \geq 0).
    \end{equation}

\subsection{Extensions to MOBO with preferences} \label{sec:preference-extension}
    
    % When the objective space is high-dimensional, finding all Pareto-optimal solutions becomes meaningless and computationally infeasible. 
    % As discussed in \citet{paria2020flexible}, we need flexible methods for MOO that can steer the sampling strategy towards regions of the Pareto front that a domain expert may be interested in.
    % While the user has the liberty to choose any preference information best suited for the application at hand, for demonstration we show one possible way. 
    % A popular way of specifying user preferences is by using bounding boxes \cite{hakanen2017using}, where the goal is to satisfy $f_{i}\in [a_i, b_i], i=1,\dots, m$.
    % Our method can naturally incorporate these preferences into the optimization process.
    In many decision-making scenarios, exploring the entire Pareto front is unnecessary \citep{paria2020flexible, hakanen2017using}. 
    Instead, domain experts often define a Region of Interest (ROI), typically specified as a hyper-rectangle bounded by $f_{i}\in [a_i, b_i], i=1,\dots, m$. 
    Our framework naturally incorporates these preferences by integrating the boundaries directly into the constraint set.
    
    \begin{figure}[htbp]
     \vskip -0.1in 
      \begin{center}
        \centerline{\includegraphics[width=0.9\columnwidth]{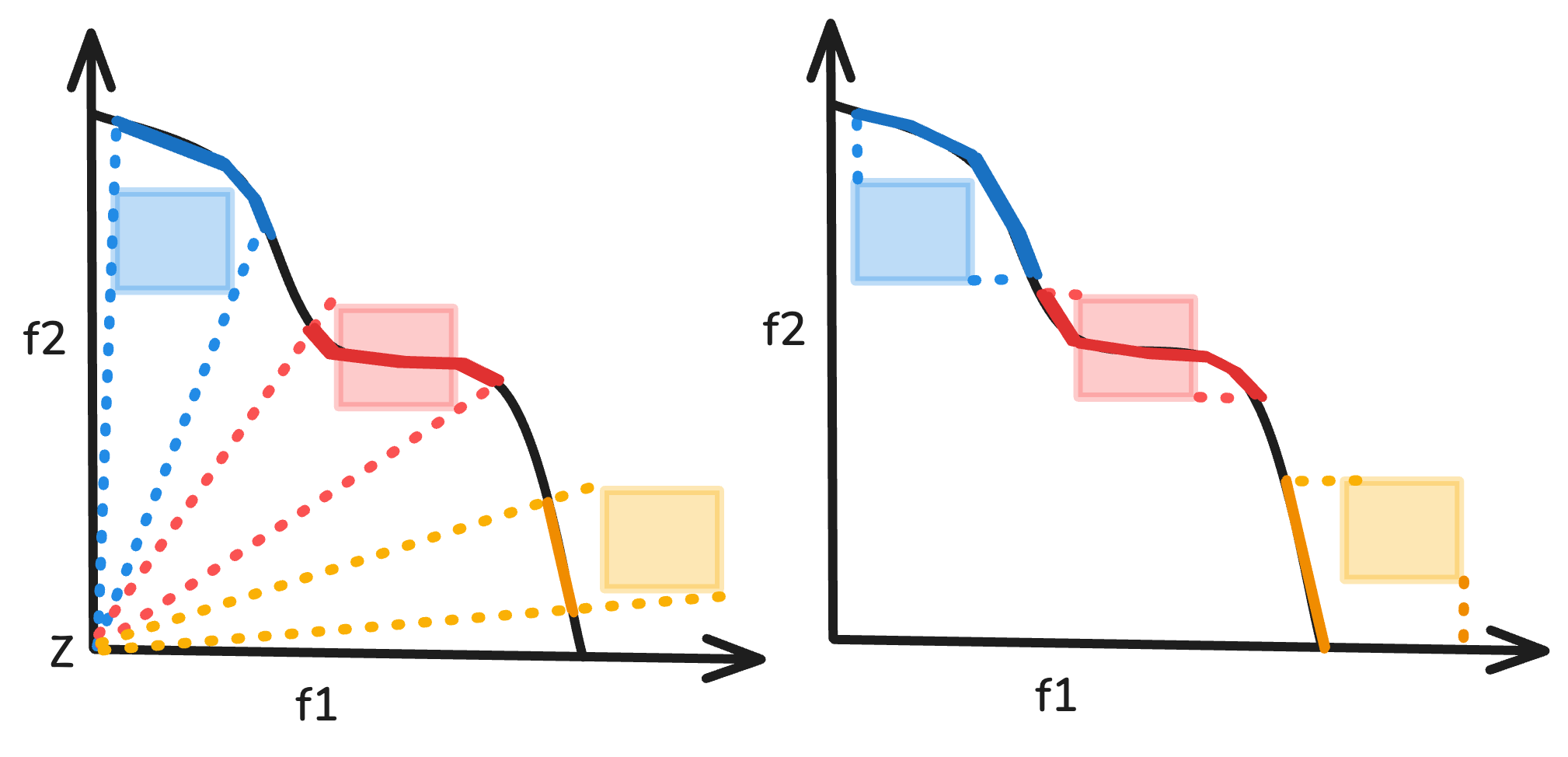}}
        \caption{\textbf{Comparison of preference handling strategies.} \textbf{Left:} Preference scalarization methods \cite{paria2020flexible} map preferred regions (shaded boxes) to preference weights dependent on a reference point $z$. \textbf{Right:} Our geometric approach operates without a reference point, directly targeting Pareto-optimal solutions that satisfy either the lower or upper bounds of the specified regions.
        }
        \label{fig:roi}
      \end{center}
       \vskip -0.2in 
    \end{figure}

    % Since decision-makers often lack prior knowledge of the true Pareto front's location, a user-defined Region of Interest (ROI) may be misaligned with the feasible space: it might be overly ambitious (located completely beyond the true front) or overly conservative (located in the dominated interior). 
    % To address this, we do not treat the ROI as a rigid binary target. 
    % Instead, we interpret the ROI as defining two distinct ``anchor'' constraints: the Lower Bounds (representing minimum acceptable criteria) and the Upper Bounds (representing aspirational ideals). 
    % Our framework seeks Pareto-optimal solutions that satisfy either of these boundary sets (\Cref{fig:roi}):
    In preference-based setting, the user-defined ROI is often misaligned with the true feasible space: it may be overly ambitious (completely beyond the true front) or overly conservative (dominated). 
    To address this issue, we do not treat the ROI as a rigid binary target. Instead, we interpret the ROI as defining two distinct anchor constraints: the lower bounds (minimum acceptable criteria) and the upper bounds (ideal limits). 
    We formulate the optimization problem to seek Pareto-optimal solutions that satisfy either of these boundary sets (\Cref{fig:roi}):
    \begin{align} \label{eq:roi_epsilon-constraint}
        \max \; &f_k(\mbf{x}) +s \sum_jf_j \\
         \text{subject to } &f_j(\mathbf{x}) \geq \varepsilon_j\; \text{for all }j=1,\dots,m, j\neq k \notag \\
         &\left\{\begin{matrix}
         (f_j(\mathbf{x}) \geq a_j\; \text{for all }j=1,\dots,m, j\neq k)\\
         \text{or} \\ 
        (f_j(\mathbf{x}) \leq b_j\; \text{for all }j=1,\dots,m, j\neq k). \\
        \end{matrix}\right. \notag
    \end{align}

    This formulation ensures robustness: if the aspirational upper bounds are unreachable, the solver anchors to the lower bounds to recover the best valid trade-offs. 
    Conversely, if the lower bounds are trivially satisfied, the upper bounds drive the search toward superior regions.

    To focus the search within the preferred region, the preference constraints are incorporated in the target identification. 
    After sampling the path in \Cref{eq:ts-path},  we solve the following cheap MOO problem with NSGA-II to obtain the sampled Pareto front in the preferred region $\widetilde{\mathcal{P}}_f^t$
    \begin{align} \label{eq:roi_nsga}
       & \widetilde{\mathcal{P}}_f^t = \max_{\mbf{x}\in \mathcal{X}}\; [\tilde{f}_1(\mbf{x}),\dots,\tilde{f}_m(\mbf{x})], \\
        &\text{subject to }\left\{\begin{matrix}
         (\tilde{f}_j(\mathbf{x}) \geq a_j\; \text{for all }j=1,\dots,m, j\neq k).\\
         \text{or} \\
        (\tilde{f}_j(\mathbf{x}) \leq b_j\; \text{for all }j=1,\dots,m, j\neq k). \\
        \end{matrix}\right. \notag
    \end{align}

    Crucially, once the target coordinate $\mathbf{Y}_c$ is identified within the preferred region, the preference information is implicitly encoded into the gap-filling $\varepsilon$-constraints ($\varepsilon_j = \widehat{\mathbf{Y}}_{c,j}$). 
    Therefore, the acquisition function remains identical to the standard version in \Cref{eq:cEI}.

    \begin{figure*}[t]
        \begin{center}
        \centerline{\includegraphics[width=\textwidth]{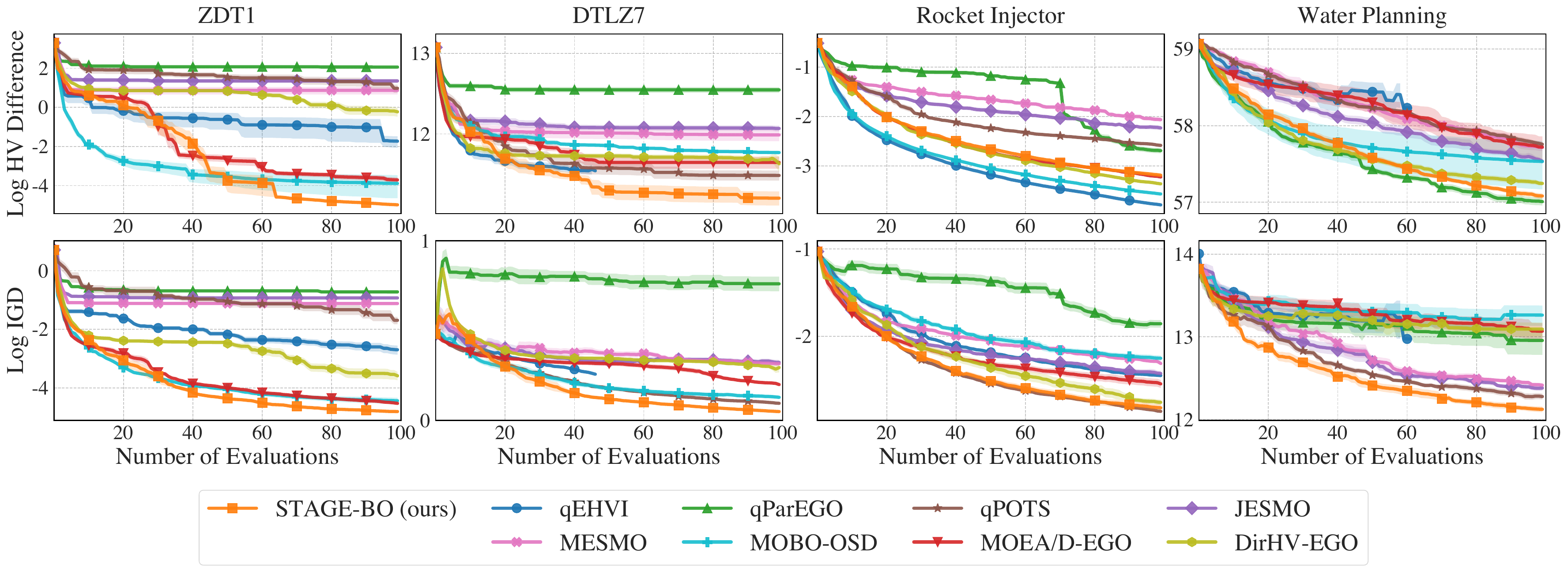}}
        \caption{
        Comparison of our method with state-of-the-art baselines on two synthetic and two real-world benchmark \textbf{MOO problems}. The first row reports hypervolume, and the second row reports IGD. Overall, our method achieves comparable or superior hypervolume relative to the baselines and consistently outperforms them in terms of IGD.
        }
        \label{fig:results_moo}
      \end{center}
        \vskip -0.2in 
    \end{figure*}

\section{Experiments} \label{sec:exps}

    We empirically evaluate our proposed method against the state-of-the-art methods on an extensive set of synthetic and real-world benchmark problems under different tasks. 
    Lastly, we present results on hyperparameter optimization for privacy-preserving machine learning, demonstrating the practical effectiveness of STAGE-BO.

\subsection{Unconstrained MOBO}
  
    \paragraph{Settings and Baselines.}
    We evaluate STAGE-BO against a comprehensive set of baselines: {qEHVI} \cite{daulton2020differentiable}, {qParEGO} \cite{knowles2006parego, daulton2020differentiable}, {JESMO} \cite{tu2022joint}, {MESMO} \cite{belakaria2019max}, {qPOTS} \cite{renganathan2025q}, {MOBO-OSD} \cite{ngomobo}, {MOEA/D-EGO} \cite{zhang2009expensive} and {DirHV-EGO} \cite{zhao2023hypervolume}.
    $q=1$.
    Detailed implementations of our method and the baselines can be found in \Cref{app:baseline}. 

    We conduct experiments on six benchmark problems.
    The number of objectives ranges from two to six, which is common in the MOBO literature. 
    For synthetic benchmark problems, we consider ZDT1 ($d=10, m=2$), ZDT2 ($d=8, m=2$) and DTLZ7 ($d=6, m=5$) with a discontinuous Pareto front. 
    For the real-world benchmark problems, we consider problems from the problem suite \cite{tanabe2020easy}: Coil compression spring design ($d=3, m=2$), Rocket injector design ($d=4, m=3$) and Water resource planning ($d=3, m=6$). 
    These problems are widely used in the MOBO literature \cite{daulton2023hypervolume, ngomobo, renganathan2025q}. 
    Details of benchmark problems can be found in \Cref{app:benchmark}.

    For the evaluation metrics, we compute the difference between the HV of the observed Pareto front and the maximum HV. 
    We also report IGD, which indicates convergence and diversity.
    We report the mean and the standard error across 10 independent runs. 

    \paragraph{Results.} 
    \Cref{fig:results_moo} summarizes the performance of STAGE-BO and all baselines.
    Note that qEHVI is evaluated for a limited number of iterations on DTLZ7 and the water planning design problem due to the prohibitively high computational cost when the number of objectives satisfies $m\geq 4$. 
    Our method consistently outperforms baselines with respect to IGD, indicating faster convergence toward the Pareto front and improved solution diversity.
    Although our method does not explicitly optimize HV, it consistently performs comparable to the best baselines, thanks to diverse coverage of the Pareto front as suggested by the low IGD.
    %the consistently low IGD values suggest good coverage and diversity of the obtained Pareto front, which in turn leads to comparable HV performance.

    % \skatt{It would be nice to say real quick what those results mentioned below are like.}
    
    Additional performance metrics, IGD+ and fill distance, are provided in \Cref{app:igd+}. 
    The full results for the ZDT2 benchmark and the Coil Compression Spring design problem are detailed in \Cref{app:more_benchmark} due to space limit. 
    Furthermore, we provide a theoretical time complexity analysis for STAGE-BO, along with a runtime comparison between STAGE-BO and the baselines in \Cref{app:time}, demonstrating that STAGE-BO maintains high efficiency in both low- and high-dimensional objective spaces.

    We also conduct ablation studies in \Cref{app:ablation} to validate the core components of our framework---our fill-distance-based constraints, the cEI acquisition function and the Thompson-sampled path. 
    Note that STAGE-BO is not specific to cEI.
    Other constrained BO acquisition functions could also be used. 
    We discuss this design choice in \Cref{app:constrained_opt}.
    The results also verify the clipping rule (Equation \ref{eq:relax}) act as a numerical stabilizer.
    Furthermore, we demonstrate that STAGE-BO is insensitive to the specific strategy used for selecting the primary optimization objective.

   \begin{figure*}[t]
      % \vskip 0.2in 
      \begin{center}
        \centerline{\includegraphics[width=\textwidth]{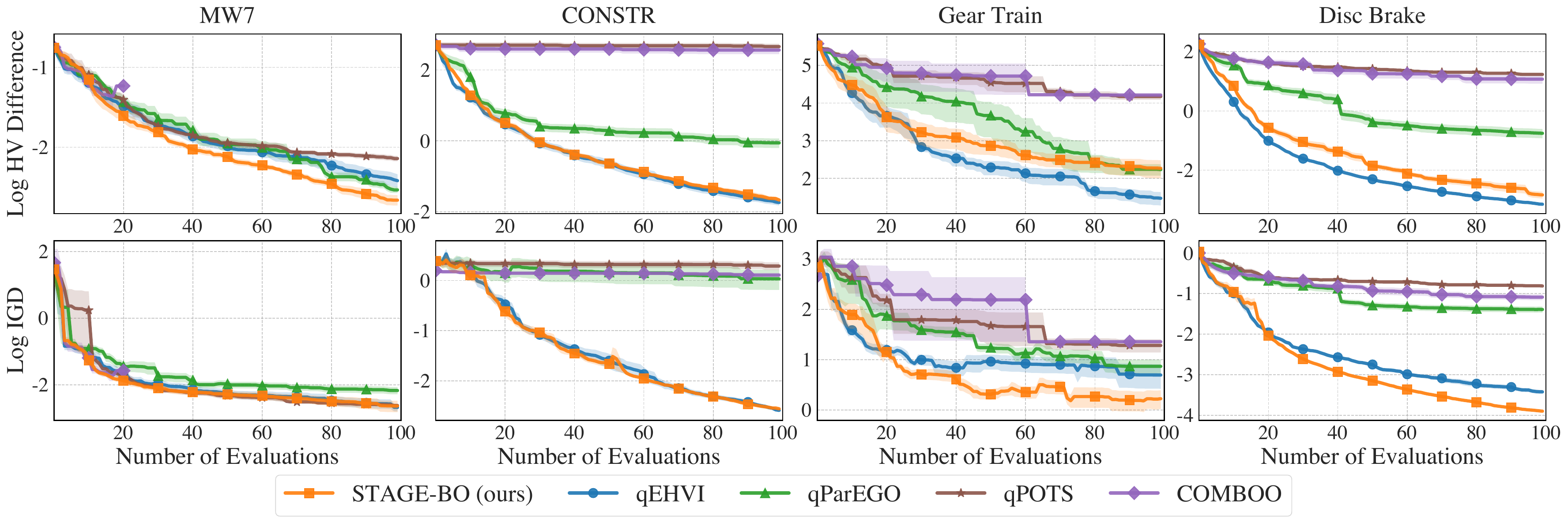}}
        \caption{
        Comparison of our method with state-of-the-art baselines on one synthetic and three real-world benchmark \textbf{constrained MOO problems}. The first row reports hypervolume, and the second row reports IGD. Overall, our method achieves comparable or superior hypervolume relative to the baselines and consistently outperforms them in terms of IGD.
        }
        \label{fig:results_con_moo}
      \end{center}
      \vskip -0.2in
    \end{figure*}
\subsection{Constrained MOBO} \label{subsec:exps_mobo_and_con_mobo}
    \paragraph{Settings and Baselines.}
    We evaluate STAGE-BO against a set of baselines: {qEHVI}, {qParEGO}, {qPOTS}, {COMBOO} \cite{li2025constrained}.
    Detailed implementations of our method and the baselines can be found in \Cref{app:baseline}. 

    % \skatt{These paragraphs take up quite a bit of space. We do not necessarily need to relist all these details all the time. Possible strategies: do not specify which baselines are compared (they can see this in the graphs). If you do specify, remove the citations (you have cited them before). Same for problem definitions below. Instead, we can say how / that the constraints are added to these methods. Other potential interesting things to mention is explaining why many fail on CONSTR, highlight that gear train and disc brake is (so far?) the only result where clearly we are trading off IGD vs HV (qEHVI consistently has higher HV but lower IGD than us).}

    We conduct experiments on four constrained MOO benchmark problems. 
    For synthetic benchmark problem, we consider MW7 ($d=4, m=2, c=2$).
    For the real-world benchmark problems, we consider two problems from the problem suite: \cite{tanabe2020easy}:  Disc brake design ($d = 4, m = 2, c= 4$), Gear train design ($d=4, m=2, c=1$), and one problem from \citet{garrido2020parallel}: CONSTR ($ d=2, m=2, c=2$).
    Details of the benchmark problems can be found in \Cref{app:benchmark}.

    \paragraph{Results.} 
    \Cref{fig:results_con_moo} shows the performance of all methods.
    Note that COMBOO is evaluated with a limited number of function evaluations on MW7, as it terminates early and returns the observed solutions when the UCB-based estimates of all constraints indicate infeasibility.
    STAGE-BO consistently outperforms state-of-the-art methods in terms of IGD, demonstrating both faster convergence and superior solution diversity.
    While our framework does not explicitly optimize for HV, its ability to ensure uniform coverage of the Pareto front inherently leads to HV performance that is competitive with, or superior to baselines.
    The results on Gear Train and Disc Brake indicate a trade-off between IGD and HV, as no single method achieves the best performance in both metrics.
    More metrics including the feasible evaluation ratio, IGD+, and fill distance are reported in \Cref{app:igd+}

\subsection{Preference-Aware MOBO} \label{subsec:exps_roi_mobo}

    \paragraph{Settings and Baselines.}
    We evaluate STAGE-BO against a set of baselines: {qEHVI}, {qParEGO}, qPOTS, {TSTCH} \cite{paria2020flexible}.
    Detailed implementations of our method and the baselines can be found in \Cref{app:baseline}. 

    \begin{figure*}[t]
      \begin{center}
        \centerline{\includegraphics[width=\textwidth]{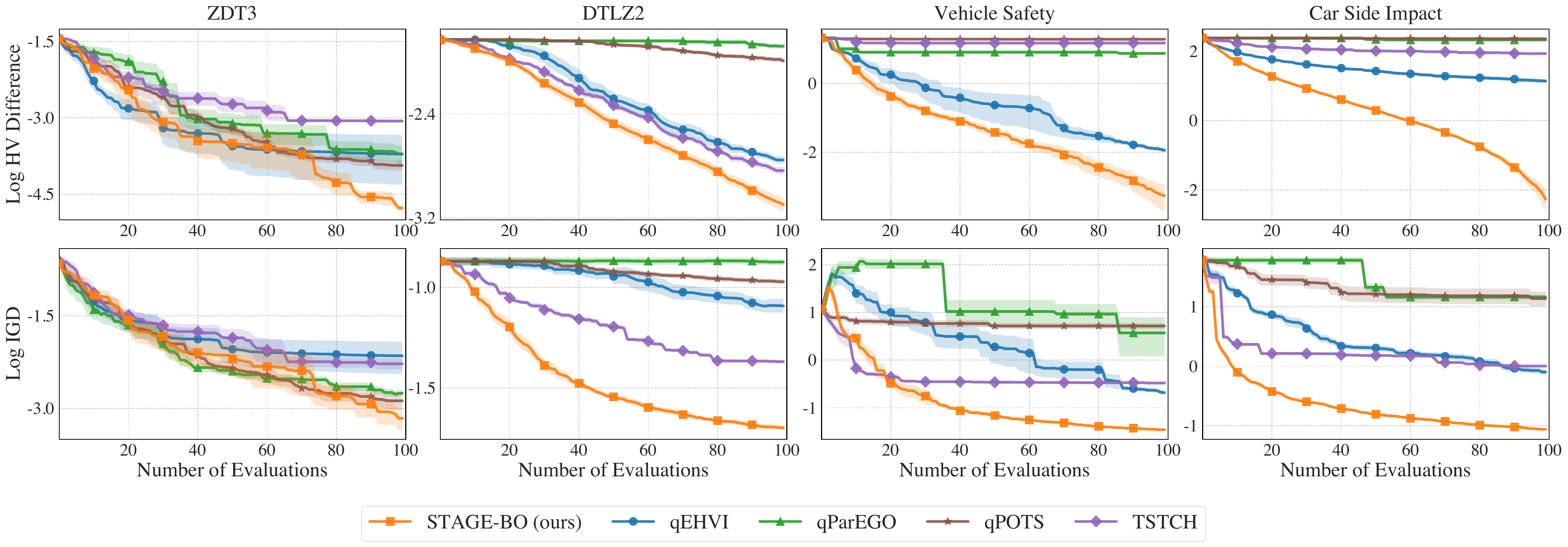}}
        \caption{
       Comparison of our method with state-of-the-art baselines on two synthetic and two real-world benchmark MOO problems with preferred regions. The first row reports hypervolume, and the second row reports IGD. Overall, our method achieves superior hypervolume relative to the baselines and consistently outperforms them in terms of IGD.
        } 
        \label{fig:results_roi_moo}
      \end{center}
      \vskip -0.2in
    \end{figure*}

    We conduct experiments on two synthetic benchmark problems:
    ZDT3 ($d=2, m=2$) with the preferred region in $([-0.7, -0.6], [-0.2, -0.4])$,
    DTLZ2 ($d=6, m=5$) with the preferred region in $[-0.4, -0.4, -0.4, -0.4, -0.4]$, $[-0.2, -0.2, -0.2, -0.2, -0.2])$,
    and two real-world benchmark problems:
    VehicleSafety Problem ($d = 5, m = 3$) with the region in $([-1680, -7, -0.5], [-1675, -6, -0.3])$, 
    CarSideImpact ($d=7, m=4$) with the region in $([-20, -4.5, -10, -7], [-15, -4, -5, -6])$. 
    Details of the benchmark problems can be found in \Cref{app:benchmark}.

    % \skatt{do we need to say something about the metrics here? In preference-aware, we do not solely care about HV and IGD, right? We care about exploring the spaces within the preference boxes? Like, if I run a preference un-aware method here, it would perform ``better'' than these current results according to these metrics?}

    \paragraph{Results.} 
    \Cref{fig:results_roi_moo} shows the performance of all methods. 
    Our method consistently performs better than competing approaches in terms of HV and IGD, indicating faster convergence toward the preferred region and improved solution diversity.
    More metrics, IGD+ and fill distance, are detailed in \cref{app:igd+}.
    We also vary the bounds of ROI in \Cref{fig:app_preference_regions,fig:region_1,fig:region_2,fig:region_3} to demonstrate that STAGE-BO is agnostic to the location of preferred regions.

\subsection{Real-world application} \label{subse:exps_real_world}

    % \skatt{We can shorten the description of the problem.}
    
    We additionally evaluate STAGE-BO on a multi-objective hyperparameter optimization task for privacy-preserving machine learning. 
    Differential privacy \cite{dwork2006calibrating} provides a formal guarantee that limits how much information about any individual training example can be inferred from the learned model, but stronger privacy protection typically requires injecting more noise during training, which can reduce predictive accuracy. 
    This creates a natural trade-off between privacy and utility, making the problem well suited to multi-objective optimization. 
    Specifically, we train a logistic regression model with differentially private stochastic gradient descent (DP-SGD) \cite{abadi2016deep} on the Dutch dataset \cite{van2000integrating}, and tune five hyperparameters controlling the training procedure.  
    The hyperparameter ranges are listed in \Cref{tab:hypers}.
    \begin{table}[h]
    \centering
    \caption{Hyperparameter search ranges.}
    \begin{tabular}{l c }
        \toprule
        \textbf{Hyperparameter} & \textbf{Range} \\
        \midrule
        Batch size & $[8,512]$  \\
        Learning rate & $[5e-4, 5e-2]$ \\
        Clipping threshold & $[0.1,4]$  \\
        Epochs& $[1,64]$  \\
        Noise level & $[1, 2560]$ \\
        \bottomrule
    \end{tabular}
    \label{tab:hypers}
\end{table}

\begin{figure}[t]
      \begin{center}
        \centerline{\includegraphics[width=0.5\textwidth]{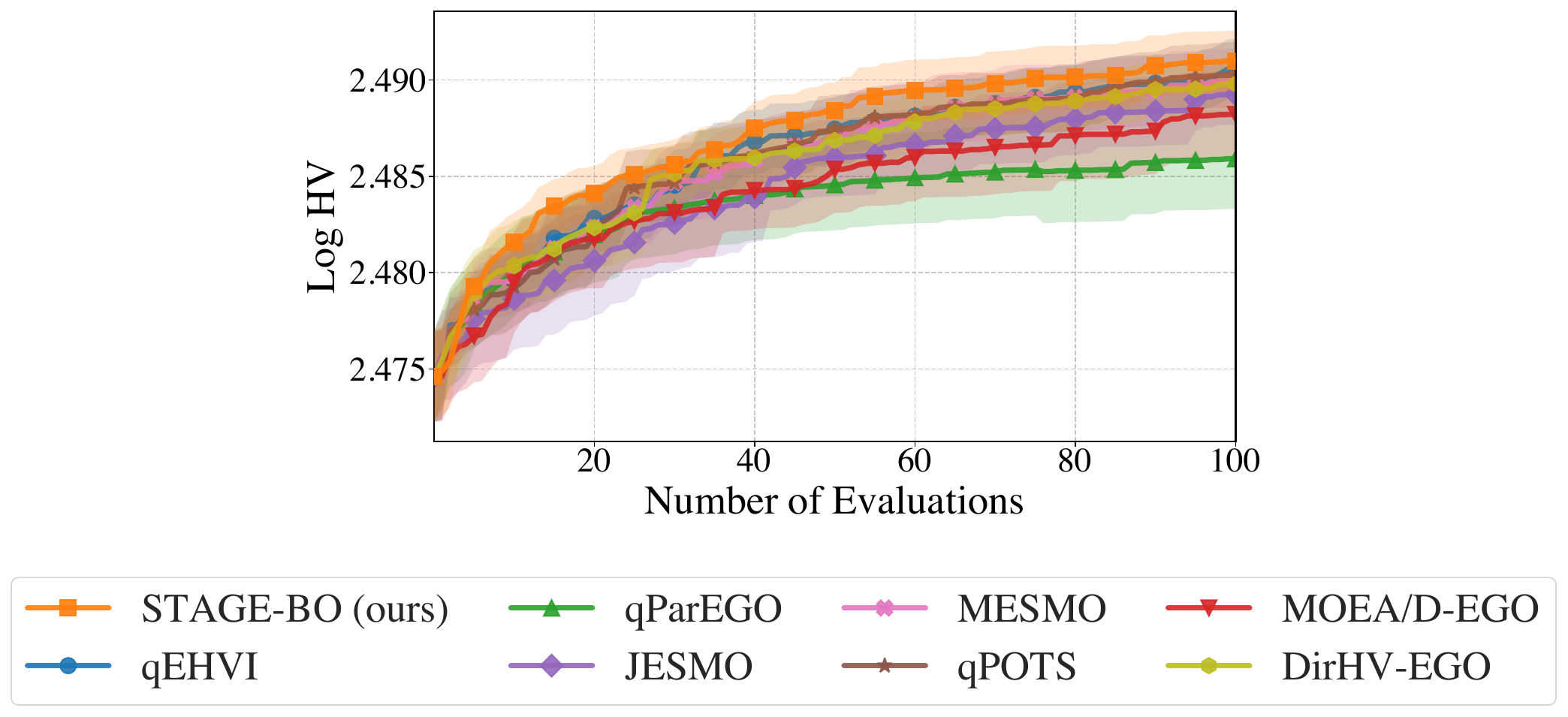}}
        \caption{
       Results on a real-world multi-objective hyperparameter optimization task for privacy-preserving machine learning. Since the true Pareto front is unknown, performance is measured by the HV of the observed non-dominated solutions. STAGE-BO achieves the highest HV throughout the optimization process.
        } 
        \label{fig:dp_mobo}
      \end{center}
      \vskip -0.3in
    \end{figure}

    We consider two competing objectives: model utility and privacy. 
    Because the true Pareto front is unknown in this real-world setting, we evaluate methods using the HV of the observed non-dominated solutions. 
    As shown in \Cref{fig:dp_mobo}, STAGE-BO achieves the strongest overall HV performance, demonstrating that the proposed framework remains effective on a practically relevant privacy–utility trade-off beyond the synthetic and engineering benchmarks.
    
% \skatt{I find the inconsistency between reporting HV and $log$ HV confusing. In particular in~\cref{fig:dp_mobo} I have difficulties: a small increase on the right side of the graph means a significant increase in real HV, correct? Also, are the standard error bounds once or two times the standard error? Lastly, in log space, the error bounds should not look symmetrical, so I am worried that these are wrong (in all figures that report log values on the y-axis).}

\section {Conclusion} \label{sec: conclusion}

    We have proposed STAGE-BO that bypasses the prohibitive computational costs of hypervolume-based methods by targeting the Pareto front through explicit geometric gap-filling. 
    By reformulating objective trade-offs as adaptive $\varepsilon$-constraints and utilizing cEI, our method provides a unified approach for recovering the global Pareto front, satisfying physical constraints, or focusing on specific regions of interest. 
    The flexibility and scalability of STAGE-BO make it a robust tool for high-stakes optimization in design and experimental sciences, where balancing diversity, efficiency, and expert preferences is paramount.
    For example, STAGE-BO could be potentially useful for future interactive MOO methods, where the uses' preferences can be encoded more flexibly than weights \cite{chen2025interactive,yang2025interactive}.

    % We have proposed a flexible and sample-efficient MOBO framework that overcomes the computational costs of hypervolume-based methods. 
    % Our method, STAGE-BO, explores the Pareto front by identifying and filling geometric gaps.
    % By treating objectives as adaptive constraints, we provided a method that is capable of recovering the entire Pareto front or focusing on specific regions of interest. 
    % % The use of constrained Thompson Sampling further ensured that these challenging sub-problems were solved efficiently. 
    % These constrained sub-problems can be efficiently solved by constrained EI.
    % % Crucially, the proposed framework unifies general Pareto exploration with the handling of complex constraints and user preferences. 
    % We believe this flexibility makes our method a valuable contribution to the toolkit of practical MOBO, particularly for high-stakes applications in design and experimental sciences.

    \paragraph{Limitations}
    First, STAGE-BO depends on the quality of the sampled Pareto front; if this approximation is poor, target identification may be inaccurate.
    In our current implementation, we use NSGA-II as the inner solver, which is effective for the low-to-moderate ($m\leq 6$) objective settings considered in this paper. 
    For higher-dimensional many-objective problems, replacing NSGA-II with NSGA-III \cite{deb2013evolutionary} may be beneficial.
    Second, our current gap detection relies on observed point positions, which may be sensitive to measurement noise. 
    % Extreme measurement errors can distort the sampled Pareto front topology and lead to suboptimal constraint placement. 
    Incorporating noise-robust geometric estimates remains a promising direction for future research.
      
\newpage
\section*{Impact Statement}

This paper presents work whose goal is to advance the field of Machine
Learning. There are many potential societal consequences of our work, none
which we feel must be specifically highlighted here.

\section*{Acknowledgments}

This work was supported by the Research Council of Finland Flagship programme: Finnish Center for Artificial Intelligence FCAI and UKRI Turing AI World-Leading Researcher Fellowship, EP/W002973/1. 
This work has been performed using resources provided by the Aalto Science-IT Project from Computer Science IT, the CSC – IT Center for Science, Finland and the Finnish Computing Competence Infrastructure (FCCI).


\bibliography{icml2026}

@inproceedings{ngomobo,
  title={MOBO-OSD: Batch Multi-Objective {Bayesian} Optimization via Orthogonal Search Directions},
  author={Ngo, Lam and Ha, Huong and Chan, Jeffrey and Zhang, Hongyu},
  booktitle={The Thirty-ninth Annual Conference on Neural Information Processing Systems},
  year={2025}
}

@inproceedings{belakaria2020uncertainty,
  title={Uncertainty-aware search framework for multi-objective {Bayesian} optimization},
  author={Belakaria, Syrine and Deshwal, Aryan and Jayakodi, Nitthilan Kannappan and Doppa, Janardhan Rao},
  booktitle={Proceedings of the AAAI Conference on Artificial Intelligence},
  volume={34},
  pages={10044--10052},
  year={2020}
}

@article{daulton2020differentiable,
  title={Differentiable expected hypervolume improvement for parallel multi-objective {Bayesian} optimization},
  author={Daulton, Samuel and Balandat, Maximilian and Bakshy, Eytan},
  journal={Advances in neural information processing systems},
  volume={33},
  pages={9851--9864},
  year={2020}
}

@article{tu2022joint,
  title={Joint entropy search for multi-objective {Bayesian} optimization},
  author={Tu, Ben and Gandy, Axel and Kantas, Nikolas and Shafei, Behrang},
  journal={Advances in Neural Information Processing Systems},
  volume={35},
  pages={9922--9938},
  year={2022}
}

@inproceedings{ahmadianshalchi2024pareto,
  title={Pareto front-diverse batch multi-objective {Bayesian} optimization},
  author={Ahmadianshalchi, Alaleh and Belakaria, Syrine and Doppa, Janardhan Rao},
  booktitle={Proceedings of the AAAI Conference on Artificial Intelligence},
  volume={38},
  pages={10784--10794},
  year={2024}
}

@article{sener2018multi,
  title={Multi-task learning as multi-objective optimization},
  author={Sener, Ozan and Koltun, Vladlen},
  journal={Advances in neural information processing systems},
  volume={31},
  year={2018}
}

@article{xu2025multi,
  title={Multi-objective optimization in machine learning assisted materials design and discovery},
  author={Xu, Pengcheng and Ma, Yingying and Lu, Wencong and Li, Minjie and Zhao, Wenyue and Dai, Zhilong},
  journal={Journal of Materials Informatics},
  volume={5},
  number={2},
  pages={N--A},
  year={2025},
  publisher={OAE Publishing Inc.}
}

@article{kouritem2022multi,
  title={A multi-objective optimization design of industrial robot arms},
  author={Kouritem, Sallam A and Abouheaf, Mohammed I and Nahas, Nabil and Hassan, Mohamed},
  journal={Alexandria Engineering Journal},
  volume={61},
  number={12},
  pages={12847--12867},
  year={2022},
  publisher={Elsevier}
}

@inproceedings{paria2020flexible,
  title={A flexible framework for multi-objective {Bayesian} optimization using random scalarizations},
  author={Paria, Biswajit and Kandasamy, Kirthevasan and P{\'o}czos, Barnab{\'a}s},
  booktitle={Uncertainty in Artificial Intelligence},
  pages={766--776},
  year={2020},
  organization={PMLR}
}

@book{garnett2023bayesian,
  title={Bayesian optimization},
  author={Garnett, Roman},
  year={2023},
  publisher={Cambridge University Press}
}

@book{williams2006gaussian,
  title={Gaussian processes for machine learning},
  author={Williams, Christopher KI and Rasmussen, Carl Edward},
  volume={2},
  year={2006},
  publisher={MIT press Cambridge, MA}
}

@inproceedings{movckus1974bayesian,
  title={On {Bayesian} methods for seeking the extremum},
  author={Mo{\v{c}}kus, Jonas},
  booktitle={IFIP Technical Conference on Optimization Techniques},
  pages={400--404},
  year={1974},
  organization={Springer}
}

@inproceedings{srinivas2010gaussian,
  title={Gaussian process optimization in the bandit setting: no regret and experimental design},
  author={Srinivas, Niranjan and Krause, Andreas and Kakade, Sham and Seeger, Matthias},
  booktitle={Proceedings of the 27th International Conference on Machine Learning},
  pages={1015--1022},
  year={2010}
}

@article{thompson1933likelihood,
  title={On the likelihood that one unknown probability exceeds another in view of the evidence of two samples},
  author={Thompson, William R},
  journal={Biometrika},
  volume={25},
  number={3/4},
  pages={285--294},
  year={1933},
  publisher={JSTOR}
}

@article{knowles2006parego,
  title={{ParEGO}: A hybrid algorithm with on-line landscape approximation for expensive multiobjective optimization problems},
  author={Knowles, Joshua},
  journal={IEEE transactions on evolutionary computation},
  volume={10},
  number={1},
  pages={50--66},
  year={2006},
  publisher={IEEE}
}

@inproceedings{hernandez2016predictive,
  title={Predictive entropy search for multi-objective {Bayesian} optimization},
  author={Hern{\'a}ndez-Lobato, Daniel and Hernandez-Lobato, Jose and Shah, Amar and Adams, Ryan},
  booktitle={International conference on machine learning},
  pages={1492--1501},
  year={2016},
  organization={PMLR}
}

@article{belakaria2019max,
  title={Max-value entropy search for multi-objective {Bayesian} optimization},
  author={Belakaria, Syrine and Deshwal, Aryan and Doppa, Janardhan Rao},
  journal={Advances in neural information processing systems},
  volume={32},
  year={2019}
}

@article{das1998normal,
  title={Normal-boundary intersection: A new method for generating the Pareto surface in nonlinear multicriteria optimization problems},
  author={Das, Indraneel and Dennis, John E},
  journal={SIAM journal on optimization},
  volume={8},
  number={3},
  pages={631--657},
  year={1998},
  publisher={SIAM}
}

@article{daulton2021parallel,
  title={Parallel {Bayesian} optimization of multiple noisy objectives with expected hypervolume improvement},
  author={Daulton, Samuel and Balandat, Maximilian and Bakshy, Eytan},
  journal={Advances in neural information processing systems},
  volume={34},
  pages={2187--2200},
  year={2021}
}

@inproceedings{daulton2023hypervolume,
  title={Hypervolume knowledge gradient: a lookahead approach for multi-objective {Bayesian} optimization with partial information},
  author={Daulton, Sam and Balandat, Maximilian and Bakshy, Eytan},
  booktitle={International Conference on Machine Learning},
  pages={7167--7204},
  year={2023},
  organization={PMLR}
}

@book{nakayama2009sequential,
  title={Sequential approximate multiobjective optimization using computational intelligence},
  author={Nakayama, Hirotaka and Yun, Yeboon and Yoon, Min},
  year={2009},
  publisher={Springer Science \& Business Media}
}

@article{tanabe2020easy,
  title={An easy-to-use real-world multi-objective optimization problem suite},
  author={Tanabe, Ryoji and Ishibuchi, Hisao},
  journal={Applied Soft Computing},
  volume={89},
  pages={106078},
  year={2020},
  publisher={Elsevier}
}

@article{bradford2018efficient,
  title={Efficient multiobjective optimization employing {Gaussian} processes, spectral sampling and a genetic algorithm},
  author={Bradford, Eric and Schweidtmann, Artur M and Lapkin, Alexei},
  journal={Journal of global optimization},
  volume={71},
  number={2},
  pages={407--438},
  year={2018},
  publisher={Springer}
}

@inproceedings{suzuki2020multi,
  title={Multi-objective {Bayesian} optimization using Pareto-frontier entropy},
  author={Suzuki, Shinya and Takeno, Shion and Tamura, Tomoyuki and Shitara, Kazuki and Karasuyama, Masayuki},
  booktitle={International conference on machine learning},
  pages={9279--9288},
  year={2020},
  organization={PMLR}
}

@article{emmerich2008computation,
  title={The computation of the expected improvement in dominated hypervolume of Pareto front approximations},
  author={Emmerich, Michael and Klinkenberg, Jan-willem},
  journal={Rapport technique, Leiden University},
  volume={34},
  pages={7--3},
  year={2008}
}

@article{konakovic2020diversity,
  title={Diversity-guided multi-objective {Bayesian} optimization with batch evaluations},
  author={Konakovic Lukovic, Mina and Tian, Yunsheng and Matusik, Wojciech},
  journal={Advances in Neural Information Processing Systems},
  volume={33},
  pages={17708--17720},
  year={2020}
}

@article{balandat2020botorch,
  title={BoTorch: A framework for efficient Monte-Carlo {Bayesian} optimization},
  author={Balandat, Maximilian and Karrer, Brian and Jiang, Daniel and Daulton, Samuel and Letham, Ben and Wilson, Andrew G and Bakshy, Eytan},
  journal={Advances in neural information processing systems},
  volume={33},
  pages={21524--21538},
  year={2020}
}

@inproceedings{gelbart2014bayesian,
  title={Bayesian optimization with unknown constraints},
  author={Gelbart, Michael A and Snoek, Jasper and Adams, Ryan P},
  booktitle={30th Conference on Uncertainty in Artificial Intelligence, UAI 2014},
  pages={250--259},
  year={2014},
  organization={AUAI Press}
}

@article{fernandez2023improved,
  title={Improved max-value entropy search for multi-objective {Bayesian} optimization with constraints},
  author={Fern{\'a}ndez-S{\'a}nchez, Daniel and Garrido-Merch{\'a}n, Eduardo C and Hern{\'a}ndez-Lobato, Daniel},
  journal={Neurocomputing},
  volume={546},
  pages={126290},
  year={2023},
  publisher={Elsevier}
}

@inproceedings{wang2017max,
  title={Max-value entropy search for efficient {Bayesian} optimization},
  author={Wang, Zi and Jegelka, Stefanie},
  booktitle={International conference on machine learning},
  pages={3627--3635},
  year={2017},
  organization={PMLR}
}

@inproceedings{hakanen2017using,
  title={On using decision maker preferences with {ParEGO}},
  author={Hakanen, Jussi and Knowles, Joshua D},
  booktitle={International Conference on Evolutionary Multi-Criterion Optimization},
  pages={282--297},
  year={2017},
  organization={Springer}
}

@inproceedings{palar2018multi,
  title={Multi-objective aerodynamic design with user preference using truncated expected hypervolume improvement},
  author={Palar, Pramudita Satria and Yang, Kaifeng and Shimoyama, Koji and Emmerich, Michael and B{\"a}ck, Thomas},
  booktitle={Proceedings of the genetic and evolutionary computation conference},
  pages={1333--1340},
  year={2018}
}

@article{he2020preference,
  title={Preference-driven Kriging-based multiobjective optimization method with a novel multipoint infill criterion and application to airfoil shape design},
  author={He, Youwei and Sun, Jinju and Song, Peng and Wang, Xuesong and Usmani, Asif S},
  journal={Aerospace Science and Technology},
  volume={96},
  pages={105555},
  year={2020},
  publisher={Elsevier}
}

@article{abdolshah2019multi,
  title={Multi-objective {Bayesian} optimisation with preferences over objectives},
  author={Abdolshah, Majid and Shilton, Alistair and Rana, Santu and Gupta, Sunil and Venkatesh, Svetha},
  journal={Advances in neural information processing systems},
  volume={32},
  year={2019}
}

@inproceedings{auger2009theory,
  title={Theory of the hypervolume indicator: optimal $\mu$-distributions and the choice of the reference point},
  author={Auger, Anne and Bader, Johannes and Brockhoff, Dimo and Zitzler, Eckart},
  booktitle={Proceedings of the tenth ACM SIGEVO workshop on Foundations of genetic algorithms},
  pages={87--102},
  year={2009}
}

@inproceedings{li2025constrained,
  title={Constrained Multi-objective {Bayesian} Optimization through Optimistic Constraints Estimation},
  author={Li, Diantong and Zhang, Fengxue and Liu, Chong and Chen, Yuxin},
  booktitle={International Conference on Artificial Intelligence and Statistics},
  pages={370--378},
  year={2025},
  organization={PMLR}
}

@inproceedings{abdolshah2018expected,
  title={Expected hypervolume improvement with constraints},
  author={Abdolshah, Majid and Shilton, Alistair and Rana, Santu and Gupta, Sunil and Venkatesh, Svetha},
  booktitle={2018 24th International Conference on Pattern Recognition (ICPR)},
  pages={3238--3243},
  year={2018},
  organization={IEEE}
}

@inproceedings{renganathan2025q,
  title={qPOTS: Efficient Batch Multiobjective {Bayesian} Optimization via Pareto Optimal Thompson Sampling},
  author={Renganathan, Ashwin and Carlson, Kade},
  booktitle={International Conference on Artificial Intelligence and Statistics},
  pages={4051--4059},
  year={2025},
  organization={PMLR}
}

@book{branke2008multiobjective,
  title={Multiobjective optimization: Interactive and evolutionary approaches},
  author={Branke, J{\"u}rgen},
  volume={5252},
  year={2008},
  publisher={Springer Science \& Business Media}
}

@article{garrido2020parallel,
  title={Parallel predictive entropy search for multi-objective {Bayesian} optimization with constraints},
  author={Garrido-Merch{\'a}n, Eduardo C and Hern{\'a}ndez-Lobato, Daniel},
  journal={arXiv preprint arXiv:2004.00601},
  year={2020}
}

@inproceedings{eriksson2021scalable,
  title={Scalable constrained {Bayesian} optimization},
  author={Eriksson, David and Poloczek, Matthias},
  booktitle={International conference on artificial intelligence and statistics},
  pages={730--738},
  year={2021},
  organization={PMLR}
}

@article{gardner2018gpytorch,
  title={Gpytorch: Blackbox matrix-matrix {Gaussian} process inference with gpu acceleration},
  author={Gardner, Jacob and Pleiss, Geoff and Weinberger, Kilian Q and Bindel, David and Wilson, Andrew G},
  journal={Advances in neural information processing systems},
  volume={31},
  year={2018}
}

@article{zhang2024gliding,
  title={Gliding over the pareto front with uniform designs},
  author={Zhang, Xiaoyuan and Li, Genghui and Lin, Xi and Zhang, Yichi and Chen, Yifan and Zhang, Qingfu},
  journal={Advances in Neural Information Processing Systems},
  volume={37},
  pages={2215--2245},
  year={2024}
}

@inproceedings{ishibuchi2015modified,
  title={Modified distance calculation in generational distance and inverted generational distance},
  author={Ishibuchi, Hisao and Masuda, Hiroyuki and Tanigaki, Yuki and Nojima, Yusuke},
  booktitle={International conference on evolutionary multi-criterion optimization},
  pages={110--125},
  year={2015},
  organization={Springer}
}

@article{schonlau1998global,
  title={Global versus local search in constrained optimization of computer models},
  author={Schonlau, Matthias and Welch, William J and Jones, Donald R},
  journal={Lecture notes-monograph series},
  pages={11--25},
  year={1998},
  publisher={JSTOR}
}

@inproceedings{gardner2014bayesian,
  title={Bayesian optimization with inequality constraints.},
  author={Gardner, Jacob R and Kusner, Matt J and Xu, Zhixiang Eddie and Weinberger, Kilian Q and Cunningham, John P},
  booktitle={ICML},
  volume={2014},
  pages={937--945},
  year={2014}
}

@article{haimes1971bicriterion,
  title={On a bicriterion formulation of the problems of integrated system identification and system optimization},
  author={Haimes, Yacov},
  journal={IEEE transactions on systems, man, and cybernetics},
  volume={3},
  pages={296--297},
  year={1971},
  publisher={Institute of Electrical and Electronics Engineers (IEEE)}
}

@book{chankong2008multiobjective,
  title={Multiobjective decision making: theory and methodology},
  author={Chankong, Vira and Haimes, Yacov Y},
  year={2008},
  publisher={Courier Dover Publications}
}

@article{mavrotas2009effective,
  title={Effective implementation of the $\varepsilon$-constraint method in multi-objective mathematical programming problems},
  author={Mavrotas, George},
  journal={Applied mathematics and computation},
  volume={213},
  number={2},
  pages={455--465},
  year={2009},
  publisher={Elsevier}
}

@article{fromer2023computer,
  title={Computer-aided multi-objective optimization in small molecule discovery},
  author={Fromer, Jenna C and Coley, Connor W},
  journal={Patterns},
  volume={4},
  number={2},
  year={2023},
  publisher={Elsevier}
}

@inproceedings{gardner2019constrained,
  title={Constrained multi-objective optimization for automated machine learning},
  author={Gardner, Steven and Golovidov, Oleg and Griffin, Joshua and Koch, Patrick and Thompson, Wayne and Wujek, Brett and Xu, Yan},
  booktitle={2019 IEEE International conference on data science and advanced analytics (DSAA)},
  pages={364--373},
  year={2019},
  organization={IEEE}
}

@article{laumanns2006efficient,
  title={An efficient, adaptive parameter variation scheme for metaheuristics based on the epsilon-constraint method},
  author={Laumanns, Marco and Thiele, Lothar and Zitzler, Eckart},
  journal={European Journal of Operational Research},
  volume={169},
  number={3},
  pages={932--942},
  year={2006},
  publisher={Elsevier}
}

@article{liu2021adaptive,
  title={Adaptive $\varepsilon$-constraint multi-objective evolutionary algorithm based on decomposition and differential evolution},
  author={Liu, Bing-Jie and Bi, Xiao-Jun},
  journal={IEEE Access},
  volume={9},
  pages={17596--17609},
  year={2021},
  publisher={IEEE}
}

@inproceedings{fan2016improved,
  title={An improved epsilon constraint handling method embedded in MOEA/D for constrained multi-objective optimization problems},
  author={Fan, Zhun and Li, Hui and Wei, Caimin and Li, Wenji and Huang, Han and Cai, Xinye and Cai, Zhaoquan},
  booktitle={2016 IEEE Symposium Series on Computational Intelligence (SSCI)},
  pages={1--8},
  year={2016},
  organization={IEEE}
}

@article{deb2002fast,
  title={A fast and elitist multiobjective genetic algorithm: NSGA-II},
  author={Deb, Kalyanmoy and Pratap, Amrit and Agarwal, Sameer and Meyarivan, TAMT},
  journal={IEEE transactions on evolutionary computation},
  volume={6},
  number={2},
  pages={182--197},
  year={2002},
  publisher={Ieee}
}

@article{zhang2009expensive,
  title={Expensive multiobjective optimization by {MOEA/D} with {Gaussian} process model},
  author={Zhang, Qingfu and Liu, Wudong and Tsang, Edward and Virginas, Botond},
  journal={IEEE Transactions on Evolutionary Computation},
  volume={14},
  number={3},
  pages={456--474},
  year={2009},
  publisher={IEEE}
}

@article{zhao2023hypervolume,
  title={Hypervolume-guided decomposition for parallel expensive multiobjective optimization},
  author={Zhao, Liang and Zhang, Qingfu},
  journal={IEEE Transactions on Evolutionary Computation},
  volume={28},
  number={2},
  pages={432--444},
  year={2023},
  publisher={IEEE}
}

@inproceedings{coello2004study,
  title={A study of the parallelization of a coevolutionary multi-objective evolutionary algorithm},
  author={Coello Coello, Carlos A and Reyes Sierra, Margarita},
  booktitle={Mexican international conference on artificial intelligence},
  pages={688--697},
  year={2004},
  organization={Springer}
}

@inproceedings{zitzler1998multiobjective,
  title={Multiobjective optimization using evolutionary algorithms—a comparative case study},
  author={Zitzler, Eckart and Thiele, Lothar},
  booktitle={International conference on parallel problem solving from nature},
  pages={292--301},
  year={1998},
  organization={Springer}
}

@inproceedings{abadi2016deep,
  title={Deep learning with differential privacy},
  author={Abadi, Martin and Chu, Andy and Goodfellow, Ian and McMahan, H Brendan and Mironov, Ilya and Talwar, Kunal and Zhang, Li},
  booktitle={Proceedings of the 2016 ACM SIGSAC conference on computer and communications security},
  pages={308--318},
  year={2016}
}

@inproceedings{dwork2006calibrating,
  title={Calibrating noise to sensitivity in private data analysis},
  author={Dwork, Cynthia and McSherry, Frank and Nissim, Kobbi and Smith, Adam},
  booktitle={Theory of cryptography conference},
  pages={265--284},
  year={2006},
  organization={Springer}
}

@article{van2000integrating,
  title={Integrating administrative registers and household surveys},
  author={Van der Laan, Paul},
  journal={Netherlands Official Statistics},
  volume={15},
  number={2},
  pages={7--15},
  year={2000}
}

@article{chen2024mosh,
  title={MoSH: Modeling Multi-Objective Tradeoffs with Soft and Hard Bounds},
  author={Chen, Edward and Dullerud, Natalie and Niedermayr, Thomas and Kidd, Elizabeth and Senanayake, Ransalu and Koh, Pang Wei and Koyejo, Sanmi and Guestrin, Carlos},
  journal={arXiv preprint arXiv:2412.06154},
  year={2024}
}

@article{chen2025interactive,
  title={Interactive Multi-Objective Probabilistic Preference Learning with Soft and Hard Bounds},
  author={Chen, Edward and Truong, Sang T and Dullerud, Natalie and Koyejo, Sanmi and Guestrin, Carlos},
  journal={arXiv preprint arXiv:2506.21887},
  year={2025}
}

@article{yang2025interactive,
  title={An Interactive Framework for Finding the Optimal Trade-off in Differential Privacy},
  author={Yang, Yaohong and Rehn, Aki and Katt, Sammie and Honkela, Antti and Kaski, Samuel},
  journal={arXiv preprint arXiv:2509.04290},
  year={2025}
}

@inproceedings{wilson2020efficiently,
  title={Efficiently sampling functions from {Gaussian} process posteriors},
  author={Wilson, James and Borovitskiy, Viacheslav and Terenin, Alexander and Mostowsky, Peter and Deisenroth, Marc},
  booktitle={International conference on machine learning},
  pages={10292--10302},
  year={2020},
  organization={PMLR}
}

@article{deb2013evolutionary,
  title={An evolutionary many-objective optimization algorithm using reference-point-based nondominated sorting approach, part I: solving problems with box constraints},
  author={Deb, Kalyanmoy and Jain, Himanshu},
  journal={IEEE transactions on evolutionary computation},
  volume={18},
  number={4},
  pages={577--601},
  year={2013},
  publisher={IEEE}
}
\bibliographystyle{icml2026}

%%%%%%%%%%%%%%%%%%%%%%%%%%%%%%%%%%%%%%%%%%%%%%%%%%%%%%%%%%%%%%%%%%%%%%%%%%%%%%%
%%%%%%%%%%%%%%%%%%%%%%%%%%%%%%%%%%%%%%%%%%%%%%%%%%%%%%%%%%%%%%%%%%%%%%%%%%%%%%%
% APPENDIX
%%%%%%%%%%%%%%%%%%%%%%%%%%%%%%%%%%%%%%%%%%%%%%%%%%%%%%%%%%%%%%%%%%%%%%%%%%%%%%%
%%%%%%%%%%%%%%%%%%%%%%%%%%%%%%%%%%%%%%%%%%%%%%%%%%%%%%%%%%%%%%%%%%%%%%%%%%%%%%%
\clearpage
\newpage
\appendix
\crefalias{section}{appendix}
\crefalias{subsection}{appendix}
\Crefname{appendix}{Appendix}{Appendices}
\crefname{appendix}{appendix}{appendices}
\onecolumn

\section{Illustration of STAGE-BO} \label{app:illustration}

\begin{figure}[htbp]
      \begin{center}
        \centerline{\includegraphics[width=1.0\textwidth]{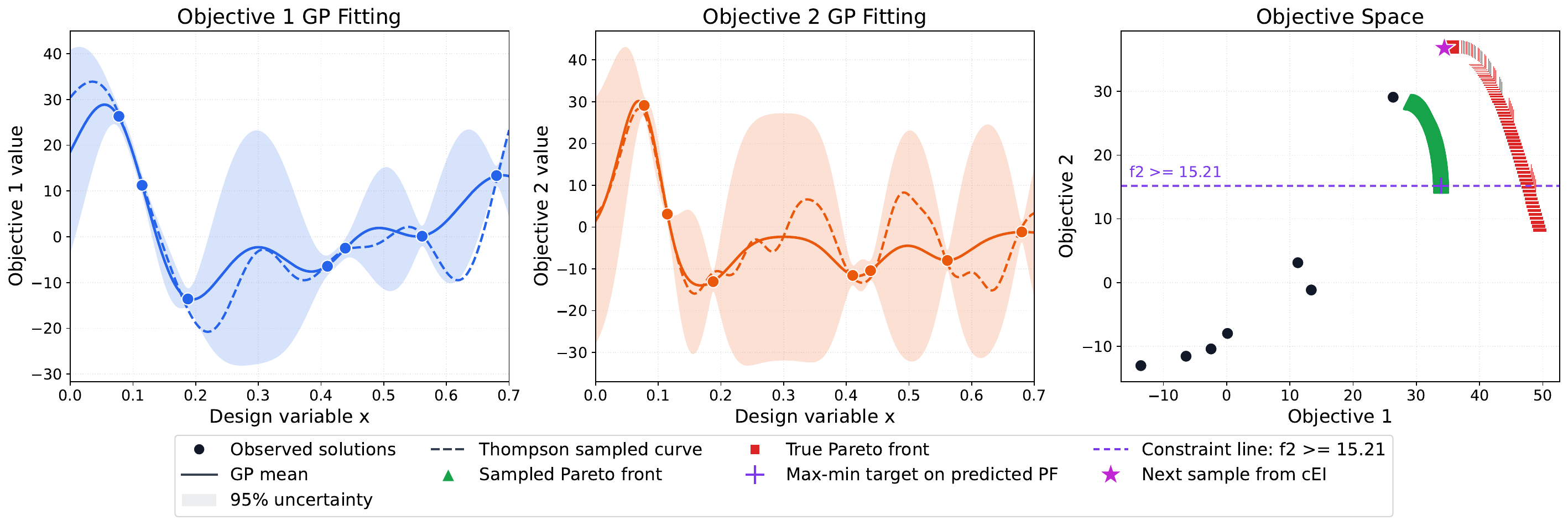}}
        \caption{
        Illustration of STAGE-BO on a two-objective problem. 
        }
        \label{fig:illustration}
      \end{center}
      \vskip -0.2in
    \end{figure}

    We visually show this process of STAGE-BO in \Cref{fig:illustration}. 
    Left and middle panels show independent GP posteriors for the two objectives. 
    A Thompson sample is drawn from the GP posterior for each objective. 
    Optimizing the sampled objectives with NSGA-II yields a sampled Pareto front (green points) in the objective space, shown in the right panel together with the true Pareto front (red points) and the currently observed objective values (black points). 
    STAGE-BO then identifies the target point (purple cross) on the sampled Pareto front by maximizing its minimum Euclidean distance to the observed front, i.e., the largest uncovered geometric gap. 
    This target is converted into an adaptive $\varepsilon$-constraint subproblem, illustrated here by the horizontal constraint line (assume we optimize $f_1$ while constraining $f_2$ at this step). 
    Constrained expected improvement (cEI) is then used to generate the next evaluation (purple star).

\section{$\varepsilon$-Constraint Multi-Objective Optimization} \label{app:epsilon-constraioned-optimization}

    The $\varepsilon$-constraint method \cite{haimes1971bicriterion, chankong2008multiobjective} is a classic multi-objective optimization strategy that transforms the original problem into a sequence of single-objective subproblems by treating all but one objective as constraints. 
    While the solution to a standard $\varepsilon$-constraint problem is guaranteed to be at least weakly Pareto optimal \cite{branke2008multiobjective}, the augmented $\varepsilon$-constraint method \cite{mavrotas2009effective}---\Cref{eq:epsilon-constraint}---introduces a small slack term to the primary objective to ensure the discovery of strictly Pareto-optimal solutions.
    
    The effectiveness of this approach depends heavily on the placement of the constraint thresholds. 
    To overcome the computational inefficiencies and poor coverage associated with fixed grids, adaptive $\varepsilon$-constraint methods \cite{laumanns2006efficient,liu2021adaptive, fan2016improved} have been proposed to iteratively refine these thresholds based on the distribution of previously discovered solutions.
    While various adaptive $\varepsilon$-constraint schemes exist, they typically rely on systematic grid refinement \cite{laumanns2006efficient} or population-based feasibility ratios \cite{fan2016improved} to adjust thresholds. 
    These heuristics are designed for settings with large evaluation budgets, such as evolutionary algorithms, and are not directly applicable to Bayesian Optimization where evaluations are prohibitively expensive. 
    In contrast, our method leverages the posterior GP belief to identify the largest geometric voids in the objective space via fill distance minimization. 
    By placing the $\varepsilon$-constraints specifically at these maxmin coordinates, we transform the $\varepsilon$-constraint method from a passive solver into a proactive, targeted acquisition strategy that ensures global diversity with a minimal number of function evaluations.

\section{Choice of Constrained Solver} \label{app:constrained_opt}

    By reformulating the MOO problem into a sequence of constrained subproblems, our framework provides the flexibility to utilize more advanced constrained Bayesian optimization methods. 
    In this work, we employ Constrained Expected Improvement (cEI) due to its robustness and mathematical simplicity in handling black-box constraints.
    
    We explicitly distinguish our approach from Trust Region (TR) methods, such as SCBO \cite{eriksson2021scalable}. 
    While TR methods are highly effective for optimization with stationary physical constraints, they are fundamentally ill-suited for our framework. 
    In STAGE-BO, the geometric $\varepsilon$-constraints are dynamic targets that shift at every iteration to target the largest under-explored voids. 
    These shifting feasible regions would frequently invalidate the internal state and local modeling of a trust region, effectively forcing it to restart and thereby negating its primary convergence benefits. 
    Consequently, global acquisition functions like cEI are more appropriate for the moving-target nature of our adaptive decomposition.

\section{Additional Experiments Results} \label{app:exps}

\subsection{Additional Evaluation Metrics} \label{app:igd+}

    Besides hypervolume and IGD, here we present more evaluation metrics.

    \paragraph{IGD+ \cite{ishibuchi2015modified}} 
    \begin{equation}\label{eq:IGD+}
        \textbf{IGD+}(\mbf{Y}_t, \mathcal{P}_f)=\frac{1}{|\mathcal{P}_f|}(\sum_{y\in\mathcal{P}_f}\min_{y'\in\mbf{Y}_t}d^+(y,y')),
    \end{equation}
    where $d^+(y,y')=\sqrt{\sum_{i=1}^m\max(y-y_i,0)^2}$.
    IGD+ is weakly Pareto-compliant \cite{ishibuchi2015modified}.
    
     \paragraph{Fill distance \cite{zhang2024gliding}} The definition of fill distance can be found in \Cref{eq:df}.
     It measures the maxmin distance between the true Pareto front and the observations. 

   \begin{figure}[htbp]
      \begin{center}
        \centerline{\includegraphics[width=\textwidth]{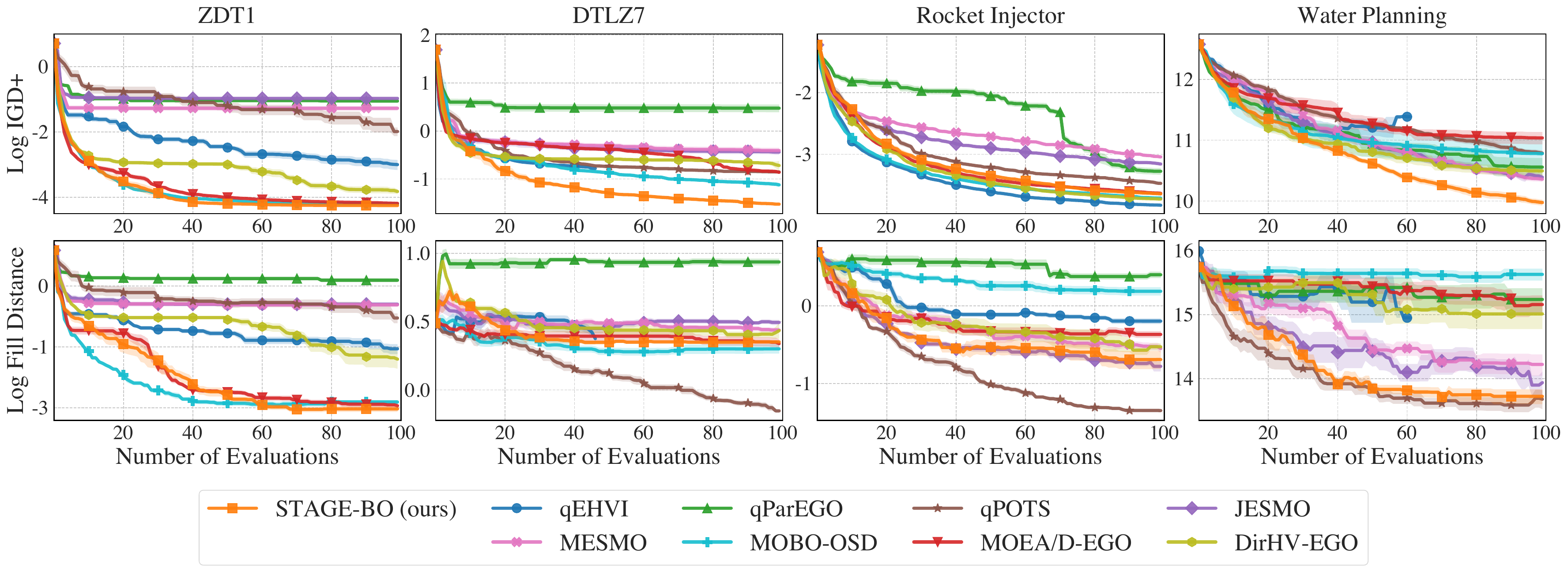}}
        \caption{
    Comparison of our method with state-of-the-art baselines on two synthetic and two real-world benchmark \textbf{MOO problems}.
The first row reports IGD+, and the second row reports fill distance. Overall, our method achieves comparable or superior performance.}
        \label{fig:others_moo}
      \end{center}
      \vskip -0.2in
    \end{figure}
    \Cref{fig:others_moo} demonstrates that STAGE-BO achieves superior performance across additional metrics. 
    While qPOTS remains a strong competitor due to its focus on maximizing diversity, it fails to achieve comprehensive coverage of the entire Pareto front; this limitation is evidenced by its significantly higher IGD+ values and lower hypervolume compared to our approach.

\begin{figure}[htbp]
      \begin{center}
        \centerline{\includegraphics[width=\textwidth]{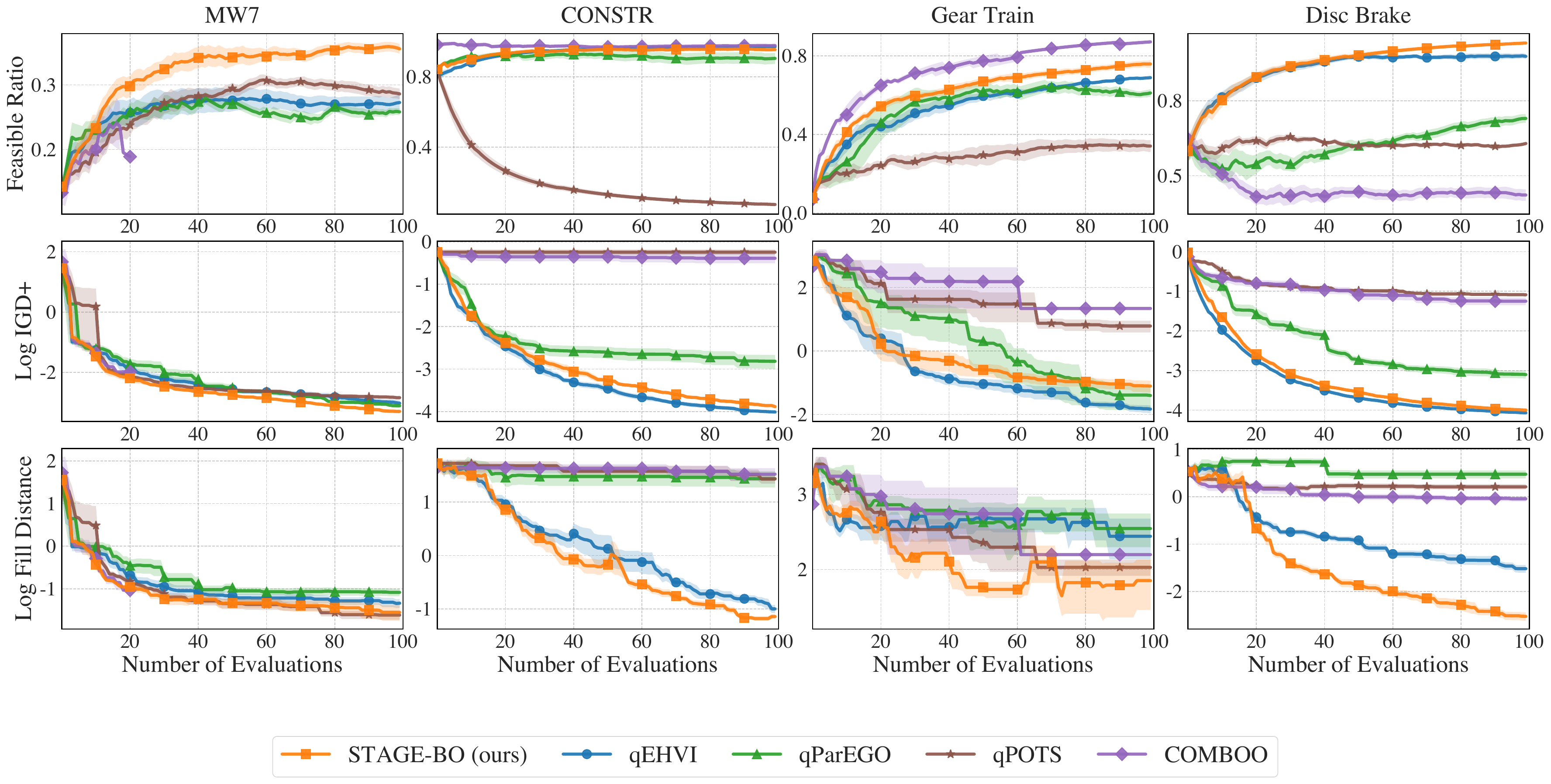}}
        \caption{
    Comparison of our method with state-of-the-art baselines on one synthetic and three real-world benchmark \textbf{constrained MOO problems}.
    The first row reports feasible ratio, the second row reports IGD+, and the last row reports fill distance. Overall, our method achieves comparable or superior performance.
        }
        \label{fig:others_con_moo}
      \end{center}
      \vskip -0.2in
    \end{figure}

    For the constrained MOO problems, in addition to IGD+ and fill distance, we report the feasibility ratio in \Cref{fig:others_con_moo}, defined as the proportion of observed points that satisfy all physical constraints throughout the optimization process.
    The consistent high feasible ratios achieved by STAGE-BO demonstrate that it can effectively estimate both the constraints and objectives to identify the feasible region and locate the constrained Pareto front.

     \begin{figure}[htbp]
      \begin{center}
        \centerline{\includegraphics[width=\textwidth]{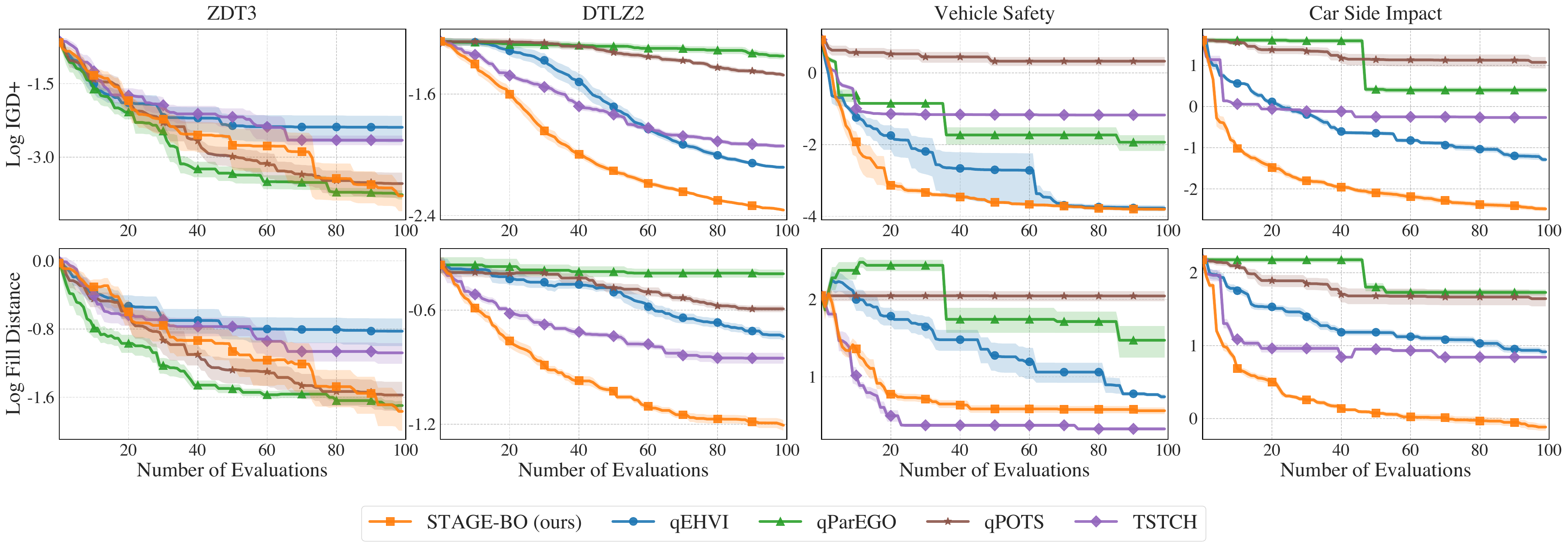}}
        \caption{
       Comparison of our method with state-of-the-art baselines on two synthetic and two real-world benchmark \textbf{MOO problems with preferred regions}. The first row reports hypervolume, and the second row reports IGD. Overall, our method  consistently outperforms baselinese in terms of IGD+ and fill distance.
        } 
        \label{fig:others_roi_moo}
      \end{center}
      \vskip -0.2in
    \end{figure}

    Lastly, we show the IGD+ and fill distance results on preference-aware MOO tasks in \Cref{fig:others_roi_moo}.
    STAGE-BO achieves superior performance across additional metrics on all datasets. 

 \begin{figure}[htbp]
      \begin{center}
        \centerline{\includegraphics[width=\textwidth]{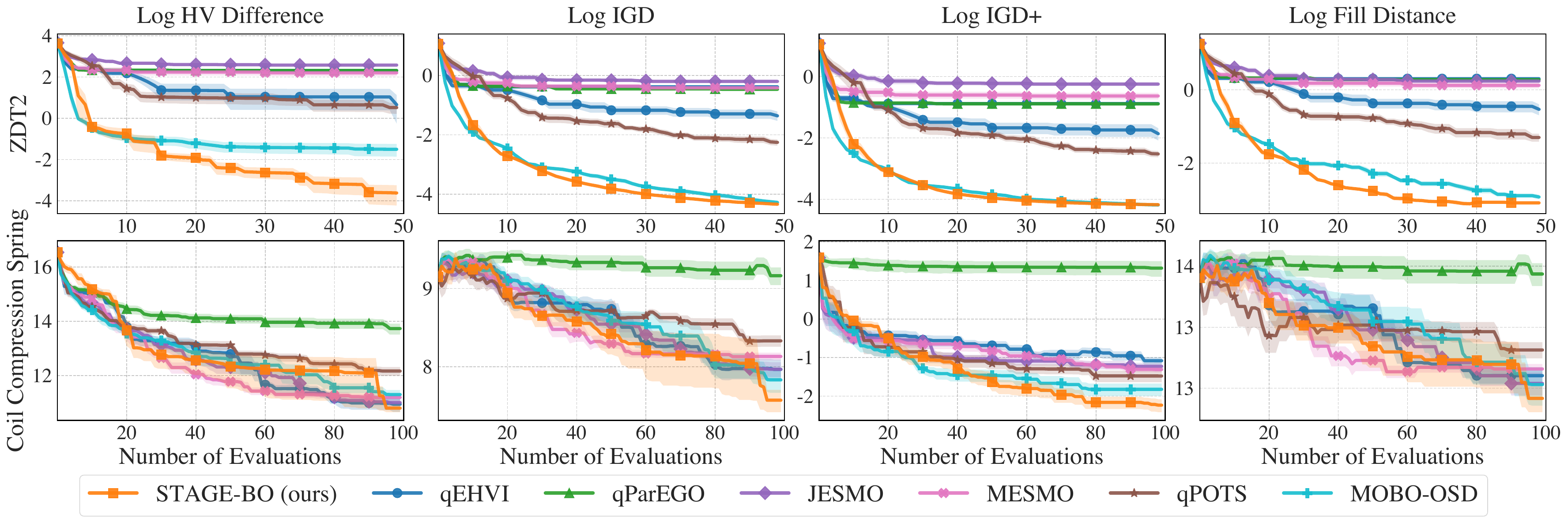}}
        \caption{
       Comparison of our method with state-of-the-art baselines on one synthetic and one real-world benchmark \textbf{MOO problems}.
       The first row reports the results for ZDT2, and the second row reports results for Coil compression spring design. Overall, our method achieves comparable or superior hypervolume relative to the baselines and consistently outperforms them in terms of IGD and fill distance.
        }
        \label{fig:app_mobo}
      \end{center}
      \vskip -0.2in
    \end{figure}
\subsection{Additional benchmark experiments} \label{app:more_benchmark}

    We present additional experimental results for the unconstrained MOO benchmarks ZDT2 ($d=8, m=2$) and the Coil Compression Spring design problem ($d=3, m=2$) in \Cref{fig:app_mobo}.

     Across both synthetic and engineering-design tasks, our method achieves superior performance in terms of HV, IGD, IGD+ and fill distance. 
     These results further validate the effectiveness of our gap-filling strategy in maintaining high solution diversity and rapid convergence toward the true Pareto front.

\subsection{Additional preferred-region experiments} \label{app:more_roi}

\begin{figure}[htbp]
      \begin{center}
        \centerline{\includegraphics[width=\textwidth]{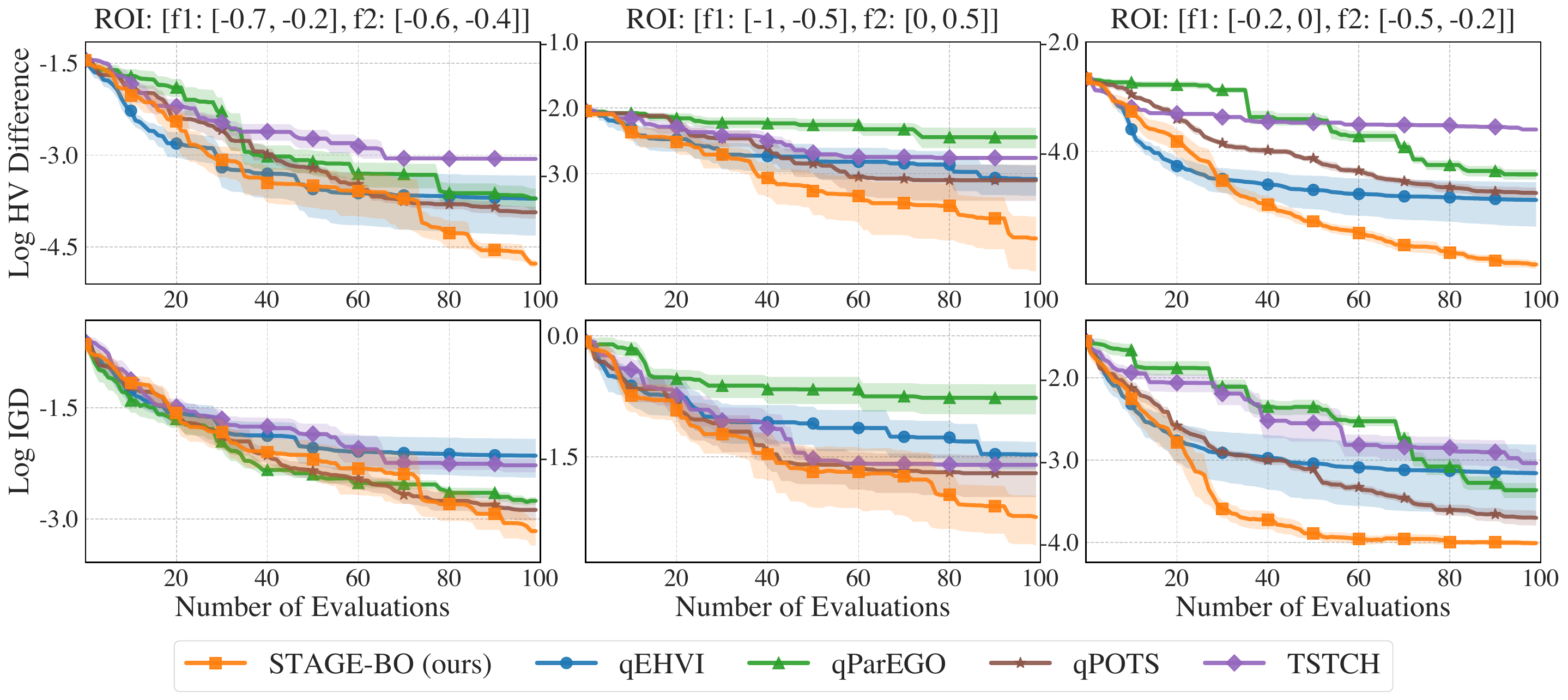}}
        \caption{
        Comparison of STAGE-BO and baseline methods on three different preferred regions of the ZDT3 benchmark. The first row reports HV difference and the seoncd row reports IGD. Across all three regions, STAGE-BO consistently achieves the lowest HV difference and IGD, demonstrating the superior performance regardless of the location of regions.
        }
        \label{fig:app_preference_regions}
      \end{center}
      \vskip -0.2in
    \end{figure}

\begin{figure}[htbp]
      \begin{center}
        \centerline{\includegraphics[width=0.8\textwidth]{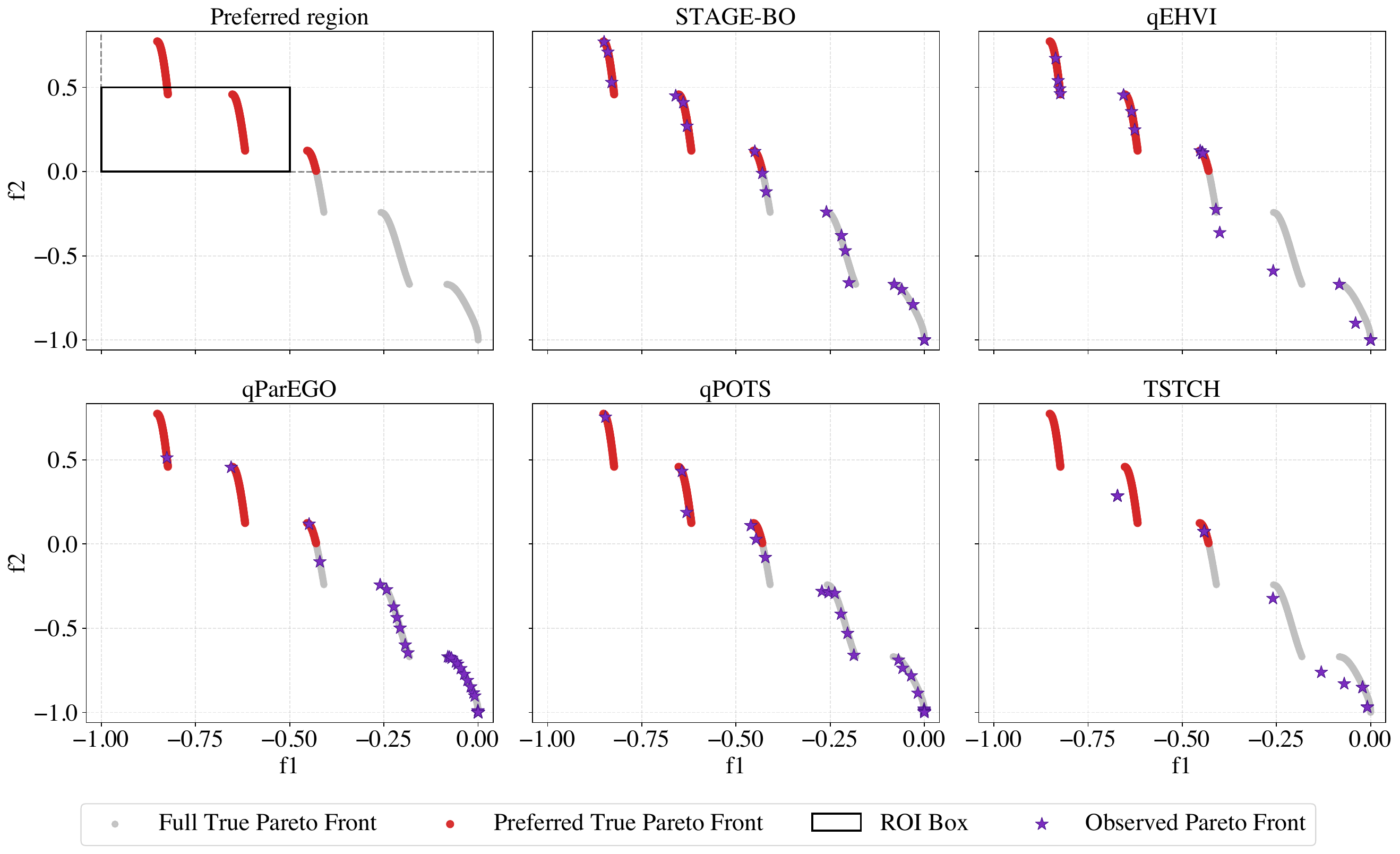}}
        \caption{
        Comparison of observed Pareto fronts (purple stars) performed by different algorithms on the ZDT3 problem. The preferred region is set on for $f_1, [-1, -0.5]$ and for $f_2, [0, 0.5]$. Grey points show the full Pareto front and red points present the preferred region. STAGE-BO can cover the preferred Pareto front more widely and uniformly.
        }
        \label{fig:region_2}
      \end{center}
      \vskip -0.2in
    \end{figure}

    We present additional experiments for the preference-based MOO benchmarks by varying the ROI---different bounds on ZDT3 ($d=2, m=2$).
    The performance results can be found in \Cref{fig:app_preference_regions}, which show that STAGE-BO performs consistently better than all baselines.
    We additionally compare the empirical observed Pareto fronts of STAGE-BO against the other baselines in \Cref{fig:region_1,fig:region_2,fig:region_3}, demonstrating it recovers the preferred Pareto front better than the baselines.

    \begin{figure}[htbp]
      \begin{center}
        \centerline{\includegraphics[width=0.8\textwidth]{figures/plot_pf_zdt3_0706.png}}
        \caption{
        Comparison of observed Pareto fronts (purple stars) performed by different algorithms on the ZDT3 problem after 100 evaluations. The preferred region is set on for $f_1, [-0.7, -0.2]$ and for $f_2, [-0.6, -0.4]$. Grey points show the full Pareto front and red points present the preferred region. STAGE-BO can cover the preferred Pareto front more widely and uniformly.        
        }
        \label{fig:region_1}
      \end{center}
      \vskip -0.2in
    \end{figure}

        \begin{figure}[htbp]
      \begin{center}
        \centerline{\includegraphics[width=0.8\textwidth]{figures/plot_pf_zdt3_0205.png}}
        \caption{
      Comparison of observed Pareto fronts (purple stars) performed by different algorithms on the ZDT3 problem. The preferred region is set on for $f_1, [-0.2, 0]$ and for $f_2, [-0.5, -0.2]$. Grey points show the full Pareto front and red points present the preferred region. STAGE-BO can cover the preferred Pareto front more widely and uniformly.
        }
        \label{fig:region_3}
      \end{center}
      \vskip -0.2in
    \end{figure}

\section{Computational Complexitye} \label{app:time}

    We provide the theoretical computational complexity analysis as follows.
    Let $N_t$ denote the number of observations at step $t$, $d$ the input dimension, $m$ the number of objectives, $R$ the number of spectral features used in Thompson sampling, $P$ the NSGA-II population size, $G$ the NSGA-II generations. 
    The per-iteration cost of STAGE-BO breaks down as follows:
    The GP training cost is well-known to scale cubically with the number of training samples, resulting in a computational complexity of $O(mN_t^3)$.
    We use Matheron construction for Thompson sampling, which results in the complexity of $O(m(dRN_t+N_t^2))$.
    The computational complexity for solving NSGA-II is $O(mGP^2+mdGP(R+N_t))$, which comprises two dominated components: the non-dominated sort $O(mGP^2)$ and point evaluation $O(mdGP(R+N_t))$.
    The runtime complexity for fill distance between P predicted Pareto front and $N_t$ observations is $O(mPN_t)$ and for cEI optimization is $O(mN_t^2)$.
    Putting the pieces together, the overall dominated term is $O(mN_t^3)$ from GP fitting. When $N_t$ is small (e.g., $N_t<200$), the NSGA-II term usually dominates. 
    Importantly, all terms are polynomial in the number of objectives and decision variables.
    By contrast, the computational complexity for qEHVI is $O(mN_t^3+mK)$, where $O(mN_t^3)$ is for the GP training and $O(mK)$ is for the box decomposition of the Pareto front. 
    The number of hyperrectangles $K$ is super-polynomial in $m$. 
    This is precisely why qEHVI becomes intractable for $m>4$.
    
\begin{figure}[htbp]
      \begin{center}
        \centerline{\includegraphics[width=\textwidth]{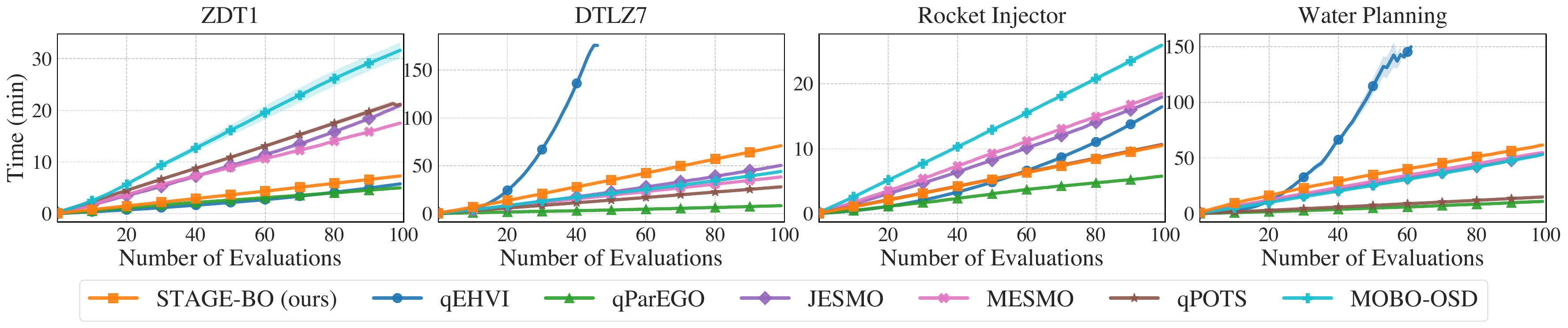}}
        \caption{
        Computation time for STAGE-BO and all baselines. Our method shows high efficiency in low dimensions when $m\leq4$ and remains computationally tractable for higher-dimensional objective spaces ($m \geq 4$).
        }
        \label{fig:time}
      \end{center}
      \vskip -0.2in
    \end{figure}

    Additionally, we compare the empirical runtime of STAGE-BO against baselines across the unconstrained MOO tasks, with results summarized in \Cref{fig:time}. 
    STAGE-BO demonstrates high efficiency for problems with fewer than four objectives ($m < 4$). 
    Furthermore, unlike many hypervolume-based methods that become computationally prohibitive as the objective space expands, our framework remains computationally tractable for higher-dimensional objective spaces ($m \geq 4$), maintaining a consistent per-iteration cost.

\section{Ablation Study} \label{app:ablation}
    We conduct ablation studies to validate the core components of STAGE-BO: effectiveness of Thompson sampling, ablation on clipping, effectiveness of targeted gap-filling, and robustness of objective selection.
    
    \subsection{Effectiveness of Thompson Sampling.} \label{app:ablation-ts}
\begin{figure}[htbp]
      \begin{center}
        \centerline{\includegraphics[width=1.0\textwidth]{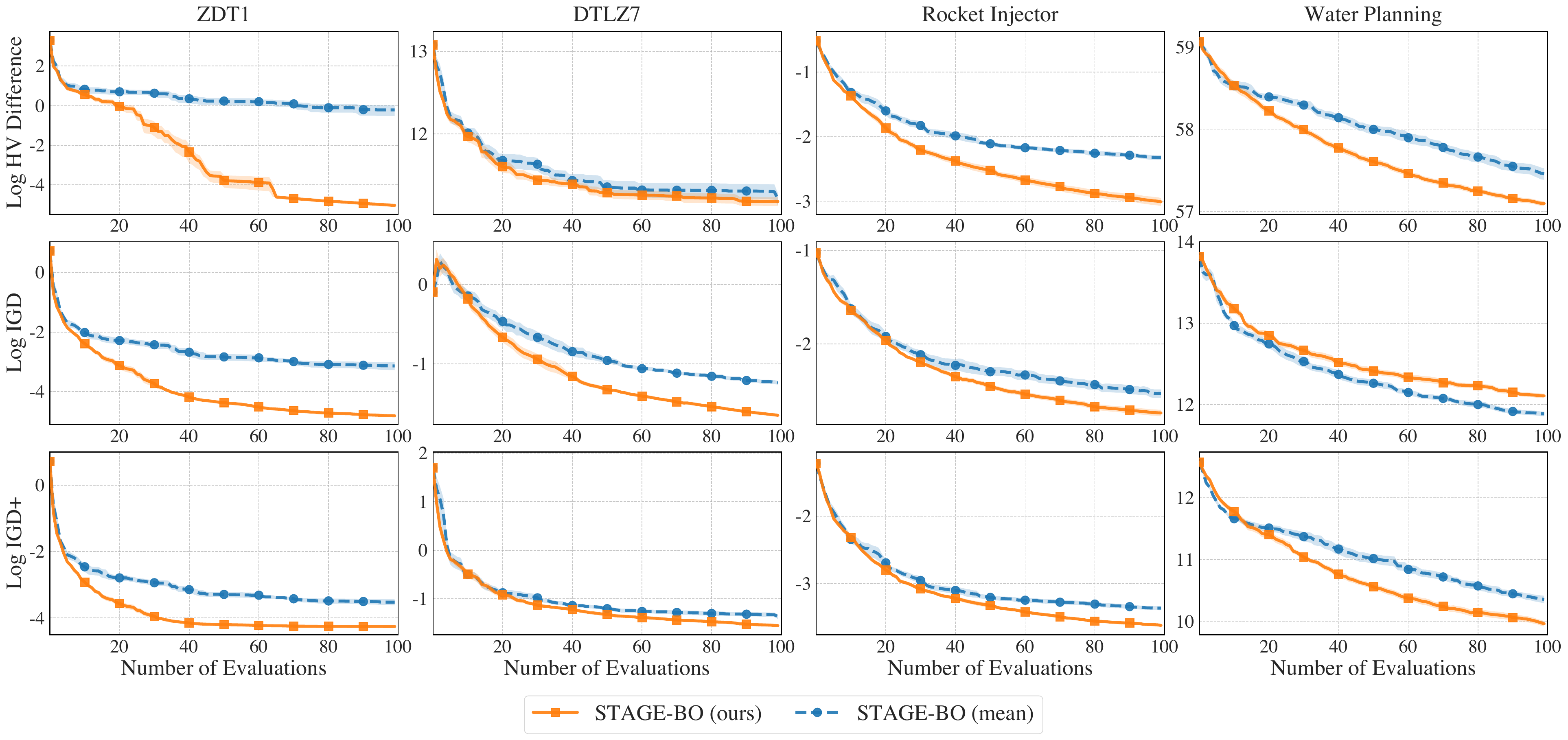}}
        \caption{
       Ablation study on replacing the single Thompson sample in STAGE-BO with the posterior mean for unconstrained MOO. 
       The three rows report log HV difference, log IGD and log IGD+, respectively. 
       The Thompson-sampling variant performs better in most cases. 
       On Water Planning, the posterior-mean variant is slightly better in IGD, while ours remains better in HV and IGD+. 
       This suggests that Thompson sampling better exploits posterior uncertainty, whereas the posterior mean is more greedy and ignores uncertainty.
        }
        \label{fig:ablation_ts}
      \end{center}
      \vskip -0.2in
    \end{figure}

    As presented in \Cref{fig:ablation_ts}, replacing the Thompson sample in STAGE-BO with the posterior mean would have the negative consequence.
    This is expected: the posterior mean is overly greedy, suppressing the uncertainty-driven variability that makes Thompson sampling effective for exploration.

    \subsection{Ablation of Clipping.} \label{app:ablation-clip}
\begin{figure}[htbp]
      \begin{center}
        \centerline{\includegraphics[width=1.0\textwidth]{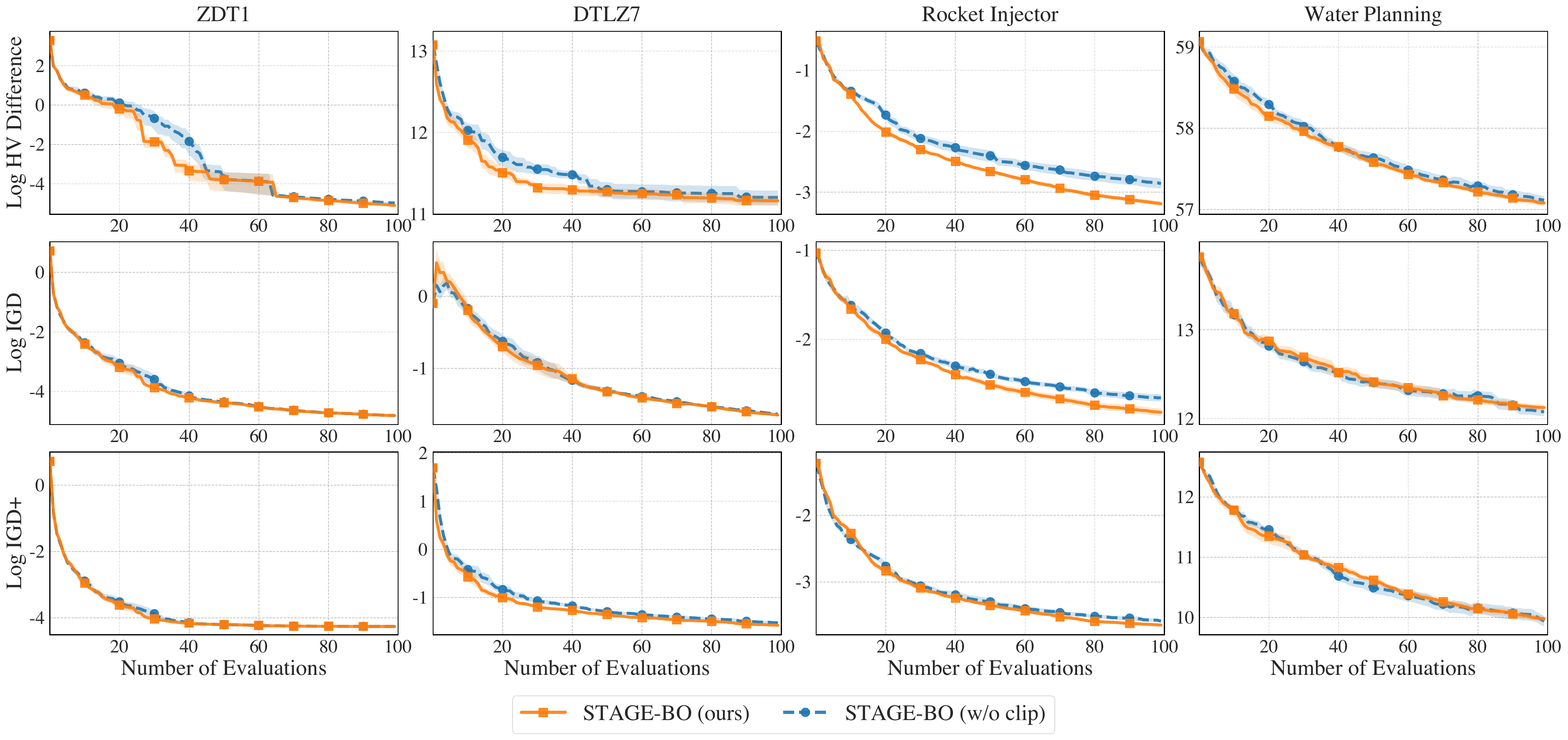}}
        \caption{
      Ablation study on clipping in STAGE-BO for unconstrained MOO. The three rows report log HV difference, log IGD and log IGD+, respectively. Clipping is designed as a numerical stabilizer and has little effect on most datasets, but consistently improves HV and IGD on Rocket Injector.
        }
        \label{fig:ablation_clip}
      \end{center}
      \vskip -0.2in
    \end{figure}

\begin{figure}[htbp]
      \begin{center}
        \centerline{\includegraphics[width=1.0\textwidth]{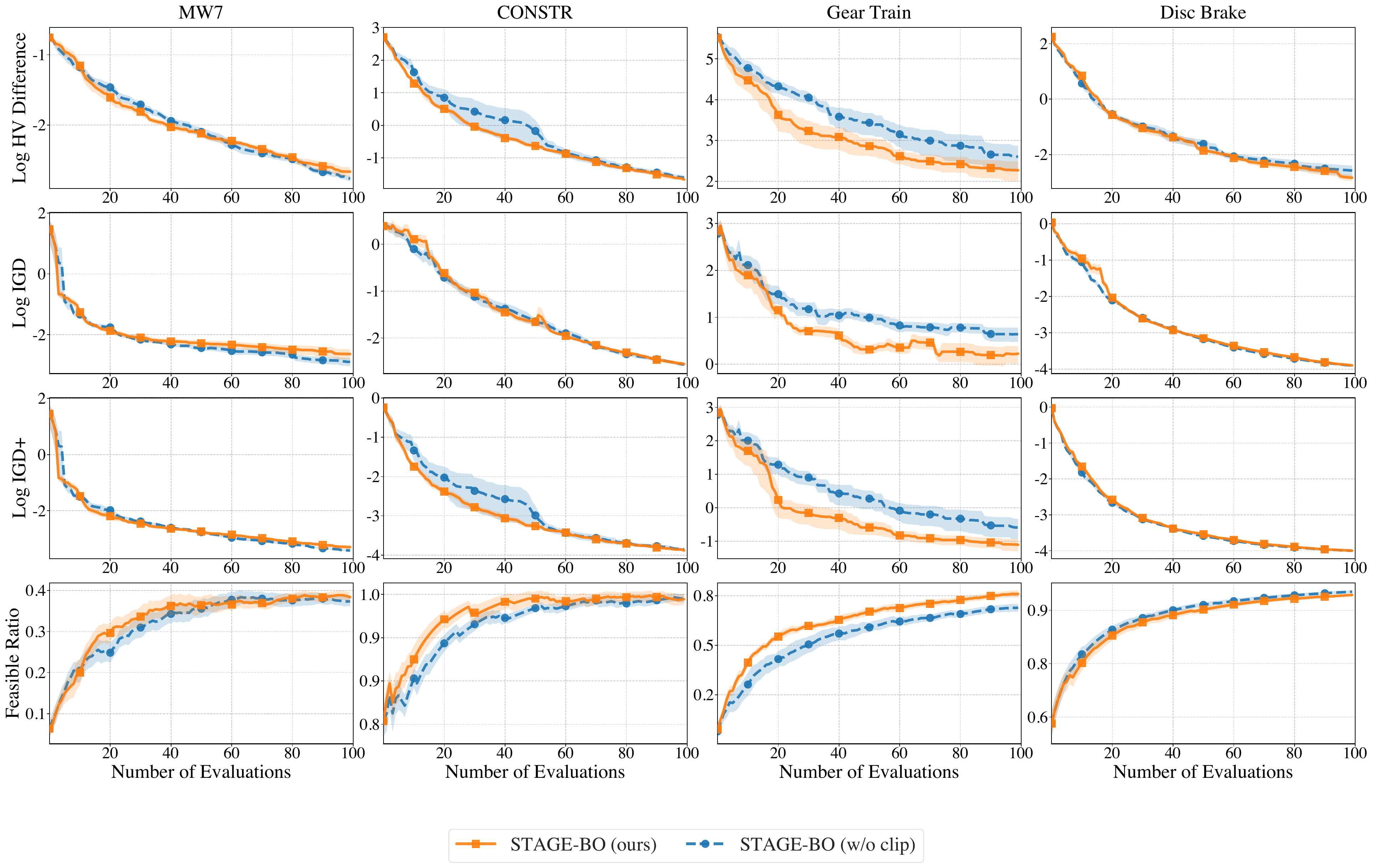}}
        \caption{
      Ablation study on clipping in STAGE-BO for constrained MOO. The four rows report log HV difference, log IGD, log IGD+ and the feasible ratio of evaluated solutions under constraints, respectively. Clipping is designed as a numerical stabilizer and has little effect on most datasets, but leads to clear gains on Gear Train.
        }
        \label{fig:ablation_con_clip}
      \end{center}
      \vskip -0.2in
    \end{figure}

\begin{figure}[htbp]
      \begin{center}
        \centerline{\includegraphics[width=1.0\textwidth]{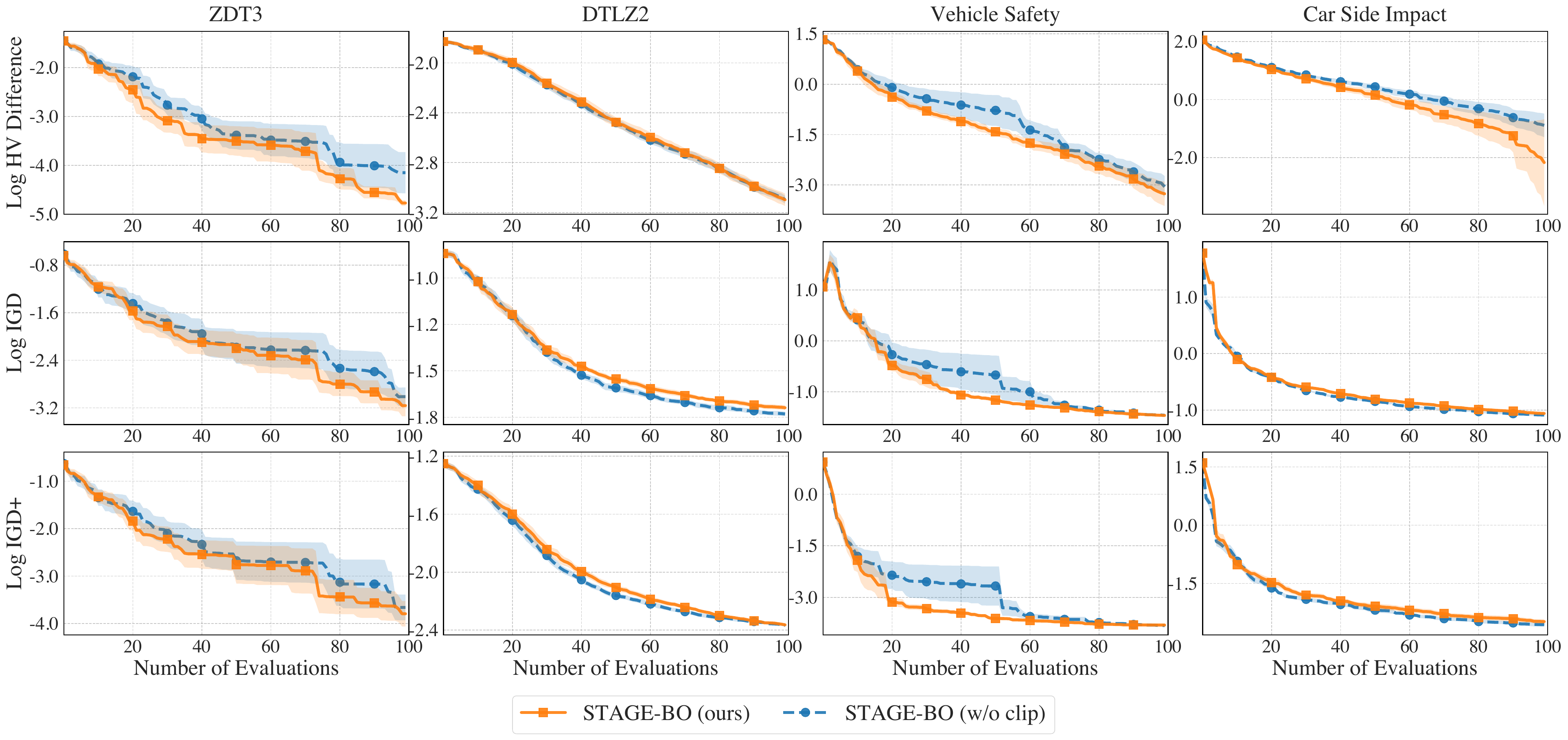}}
        \caption{
      Ablation study on clipping in STAGE-BO for preference-based MOO. The three rows report HV difference, IGD and IGD+, respectively. The clipped and unclipped variants perform similarly across all datasets, indicating that clipping acts as a numerical stabilizer.
        }
        \label{fig:ablation_preference_clip}
      \end{center}
      \vskip -0.2in
    \end{figure}

    We conducted an ablation study comparing STAGE-BO with and without clipping. 
    In practice, clipping arises primarily in early iterations when the posterior is uncertain and the sampled front overshoots the currently attainable region. 
    As the number of observations grows and the posterior concentrates, the sampled front increasingly aligns with the true front and clipping is triggered less frequently.
    Results in \Cref{fig:ablation_clip,fig:ablation_con_clip,fig:ablation_preference_clip} show that on most benchmarks the two variants perform comparably, confirming that clipping acts primarily as a stabilizer. On a subset of benchmarks, clipping leads to measurable improvements, suggesting that the larger feasible region induced by clipping benefits optimization.
    
    % We conduct an ablation study on the ZDT1 and Rocket injector design benchmark to evaluate the contribution of each component within the STAGE-BO framework. 
    % The results, illustrated in \Cref{fig:ablation}, are divided into two primary investigations: the effectiveness of our targeted acquisition and the robustness of our objective selection strategy.
    
    \subsection{Effectiveness of Targeted Gap-Filling.} \label{app:ablation-constraints}
    The left panels of \Cref{fig:ablation} compare the STAGE-BO pipeline on ZDT1 against two variants: 
    \begin{enumerate}
        \item \textbf{Direct Sampling (without cEI): }
        We bypass the acquisition optimization and directly query the maxmin target $\mathbf{Y}_c$. 
        This evaluates whether cEI provides a necessary push toward the true front beyond the raw posterior samples.
        \item \textbf{Random Constraints (No Maxmin):}
        We replace the geometric target $\mathbf{Y}_c$ with a randomly selected coordinate within the observed front to verify the necessity of explicit gap-filling via fill-distance minimization. 
        To ensure this randomly selected target defines a non-empty feasible region, we employ a Lexicographical Constraint-Setting procedure: we sequentially determine the thresholds $\varepsilon_j$ by validating feasibility for each objective in order ($f_1, f_2, \dots, f_m$), ensuring that each subsequent constraint remains reachable given the previous ones.
    \end{enumerate}  

    The results support that the combination of maxmin targeting and cEI optimization is essential for achieving superior convergence and uniform coverage.
    
    \subsection{Robustness of Objective Selection.}  \label{app:ablation-objective}
\begin{figure}[htbp]
      \begin{center}
        \centerline{\includegraphics[width=0.8\textwidth]{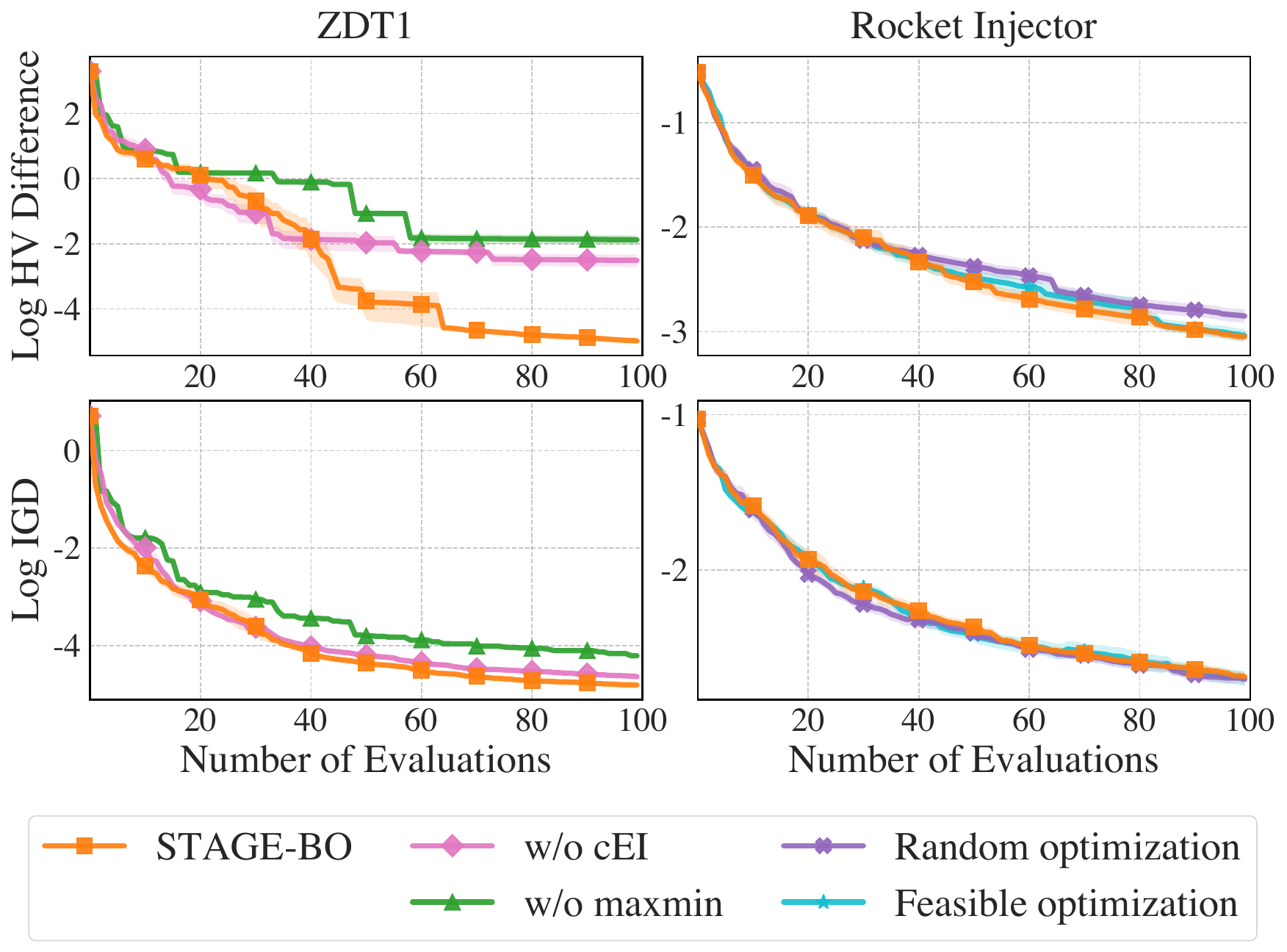}}
        \caption{
        Ablation study on different parts of our method. 
        The left panels show that cEI and computing maxmin to set the constraints are necessary. 
        The right panels show that our method is robust to the strategy of picking the objective to optimize.
        }
        \label{fig:ablation}
      \end{center}
      \vskip -0.2in
    \end{figure}

    The right panels of \Cref{fig:ablation} investigate the schedule for selecting the primary objective $f_k$.
    We compare our default round-robin schedule against two alternatives:
    \begin{enumerate}
        \item \textbf{Random Optimization:} The objective $f_k$ is chosen at random in each iteration.
        \item \textbf{Feasible Optimization:} To maximize the potential feasible region for the acquisition solver, we identify the objective $f_k$ with the minimum coordinate distance between $\mathbf{Y}_c$ and its nearest observed neighbor, thereby optimizing the dimension with the most ``crowded'' constraints.
    \end{enumerate}
    The results demonstrate that STAGE-BO is insensitive to the objective selection strategy.

\section{Baselines Implementation} \label{app:baseline}

    We implemented STAGE-BO and all baselines in Python (version 3.10). 
    The detailed implementation are as follows.

    \paragraph{STAGE-BO} For the surrogate model, we implement the GPs via GPyTorch \cite{gardner2018gpytorch} and BoTorch \cite{balandat2020botorch}.
    We employ a Matern 5/2 kernel with ARD length-scales. 
    The Gaussian likelihood is modeled with homoskedastic noise, and model hyperparameters are optimized by maximizing the Sum Marginal Log-Likelihood. 
    During fitting, a Cholesky jitter of $10^{-3}$ is applied to maintain numerical stability.
    For the Thompson sampling, we draw a sample path from each GP posterior using Matheron's rule \cite{wilson2020efficiently}, which decomposes the posterior sample into a prior spectral sample and a data-dependent update term, yielding a continuous and differentiable approximation of each objective.

     We set 300 population size and 50 max generations for the NSGA-II and use BoTorch to optimize cEI with 20 starters. 

    \paragraph{qEHVI \cite{daulton2020differentiable} } 
    We use the default hyperparameter settings from the paper and the open-sourced implementation can be found at \url{https://github.com/pytorch/botorch}.
    In the preference-aware setting, qEHVI needs to be adapted. 
    Simply setting the reference point at the corner of the ROI hypercube is not always appropriate. 
    As illustrated in \Cref{fig:roi}, the user-defined ROI may be overly ambitious and lie entirely beyond the attainable front. 
    In this case, setting the reference point at the corner of the ROI hypercube means no observed solution dominates the ROI corner, and qEHVI accumulates zero hypervolume improvement throughout the entire optimization and becomes ineffective.
    To avoid this, we adaptively set the reference point for qEHVI in a way consistent with our framework.
    We construct the qEHVI reference point from a Thompson-sampled path: we sample a posterior path over the design space, identify sampled points that fall inside the feasible preference region, and set the reference point to their coordinate-wise lower bounds. 
    If no sampled point falls inside the preferred region, we fall back to the coordinate-wise lower bounds of the full sampled path.

    \paragraph{qParEGO \cite{knowles2006parego}}
    qParEGO is a novel extension from ParEGO \cite{knowles2006parego} that is developed by \citet{daulton2020differentiable} to leverage batch setting. 
    We use the settings as follows: augmented Tchebychev scalarization \cite{nakayama2009sequential} and EI acquisition function with gradient solver. 
    We use the open-sourced implementation at \url{https://botorch.org/docs/tutorials}.

    \paragraph{MESMO \cite{wang2017max}}
    We use the default hyperparameter settings and the open-sourced implementation can be found at \url{https://github.com/pytorch/botorch}.

    \paragraph{JESMO \cite{tu2022joint}}
     We use the default hyperparameter settings and the open-sourced implementation can be found at \url{https://github.com/pytorch/botorch}.

    \paragraph{qPOTS \cite{renganathan2025q}}
    We use the default hyperparameter settings. 
    This includes the NSGA-II hyperparameter settings.
    We use the open-sourced implementation at \url{https://github.com/csdlpsu/qpots}.

    \paragraph{MOBO-OSD \cite{ngomobo}}
    We use the default hyperparameter settings from the paper. 
    This includes the number of points on approximated CHIM and the number of starting points when solving MOBO-OSD subproblem.
    We use the open-sourced implementation at \url{https://github.com/LamNgo1/mobo-osd}.

    \paragraph{COMBOO \cite{li2025constrained}}
    We use the default hyperparameter setting from the paper. 
    The open-sourced implementation can be found at \url{https://github.com/dancewithDianTong/ COMBOO}.

    \paragraph{TSTCH \cite{paria2020flexible}} 
    We use the settings as follows: augmented Tchebychev scalarization \cite{nakayama2009sequential} and Thompson sampling acquisition function.
    We implement this with BoTorch \cite{balandat2020botorch}.

    \paragraph{MOEA/D-EGO \cite{zhang2009expensive}}
     We use the default hyperparameter setting from the paper. 
    The open-sourced implementation can be found at \url{https://github.com/mobo-d/MOEAD-EGOO}.

    \paragraph{DirHV-EGO \cite{zhao2023hypervolume}} 
    We use the default hyperparameter setting from the paper. 
    The open-sourced implementation can be found at \url{https://github.com/mobo-d/DirHV-EGO}.

\section{Benchmark Problems} \label{app:benchmark}

Here we present the benchmark problems used in unconstrained MOO, constrained MOO and preference-aware MOO. 

    \begin{table}[h]
    \centering
    \caption{Benchmark problem settings and reference points in \textbf{unconstrained multi-objective problems}.}
    \begin{tabular}{l c c c c}
        \toprule
        \textbf{Problem} & \textbf{d} & \textbf{m} & \textbf{Reference Point} \\
        \midrule
        ZDT1 & 10 & 2 & $(-11.0, -11.0)$ \\
        ZDT2 & 8 & 2 &  $(-11.0, -11.0)$ \\
        DTLZ7 &6 & 5 &  $(-1.1, -1.1, -1.1, -1.1, -1.1)$ \\
        Coil Compression Spring& 3 & 2 &  $(-133.65, -9056129.08)$ \\
        Rocket Injector& 4 & 3 &  $(-0.96, -1.11, -1.08)$ \\
        Water Planning & 3 & 6 &  $(-84348.75, -1460.57, -3101483.5, -12442799.73, -67029.71, -1.59)$ \\
        \bottomrule
    \end{tabular}
    \label{tab:problem_settings_moo}
\end{table}

     \begin{table}[h]
    \centering
    \caption{Benchmark problem settings and reference points in \textbf{constrained multi-objective problems}.}
    \begin{tabular}{l c c c c}
        \toprule
        \textbf{Problem} & \textbf{d} & \textbf{m} & \textbf{C} & \textbf{Reference Point} \\
        \midrule  
        MW7& 4 & 2 &2 &$(-1.2, -1.2)$ \\
        CONSTR & 2 & 2& 2 & $(-10.0, -10.0)$\\
        Disc brake design& 4 & 2 & 4&$(-7.58, -7.0)$ \\
        Gear train design  & 4 & 2 & 1 &$(-7.4, -64.1)$ \\
        \bottomrule
    \end{tabular}
    \label{tab:problem_settings_cmoo}
\end{table}

 \begin{table}[h]
    \centering
    \caption{Benchmark problem settings and reference points in \textbf{preference-aware multi-objective problems}.}
    \begin{tabular}{l c c c c}
        \toprule
        \textbf{Problem} & \textbf{d} & \textbf{m}  & \textbf{Preferred Region} \\
        \midrule  
        ZDT3& 2 & 2  &$(-1,-1)$ \\
        DTLZ2 & 6 & 5  & $(-0.8442, -0.8999, -0.8358, -0.8710, -0.8553)$\\
        VehicleSafety& 5 & 3  &$(-1680, -7.0, -0.26)$ \\
        CarSideImpact  & 7 & 4   &$(-23.01,  -4.43, -13.09,  -9.47)$ \\
        \bottomrule
    \end{tabular}
    \label{tab:problem_settings_roimoo}
\end{table}

%%%%%%%%%%%%%%%%%%%%%%%%%%%%%%%%%%%%%%%%%%%%%%%%%%%%%%%%%%%%%%%%%%%%%%%%%%%%%%%
%%%%%%%%%%%%%%%%%%%%%%%%%%%%%%%%%%%%%%%%%%%%%%%%%%%%%%%%%%%%%%%%%%%%%%%%%%%%%%%

\end{document}